\documentclass{article}

\usepackage{microtype}
\usepackage{graphicx}
\usepackage{subcaption}
\usepackage{booktabs} 
\usepackage{xspace}

\usepackage{hyperref}

\usepackage[accepted]{icml2026}

\usepackage{amsmath}
\usepackage{amssymb}
\usepackage{mathtools}
\usepackage{amsthm}

\usepackage{tcolorbox}
\usepackage[table]{xcolor}

\definecolor{myblue}{rgb}{0.88,0.98,1}
\definecolor{mygreen}{rgb}{0.92, 1.0, 0.92}
\definecolor{myred}{rgb}{1, 0.9, 0.9}
\definecolor{mydarkred}{rgb}{0.8,0.02,0.02}
\definecolor{mydarkorange}{rgb}{0.40,0.2,0.02}
\definecolor{mypurple}{RGB}{239,229,253}
\definecolor{mygold}{rgb}{0.75,0.6,0.12}
\definecolor{mydarkgray}{rgb}{0.66, 0.66, 0.66}
\definecolor{mydarkgreen}{rgb}{0.02,0.6,0.02}
\definecolor{mygray}{gray}{0.9}

\usepackage[capitalize,noabbrev]{cleveref}

\theoremstyle{plain}

\theoremstyle{definition}

\theoremstyle{remark}

\usepackage[textsize=tiny]{todonotes}

\newcommand{\boldres}[1]{{\textbf{#1}}}
\newcommand{\secondres}[1]{{\underline{#1}}}

\icmltitlerunning{Thoth: Mid-Training Bridges LLMs to Time Series Understanding}

\newcommand{\icon}{\raisebox{0pt}{\includegraphics[width=1.0em]{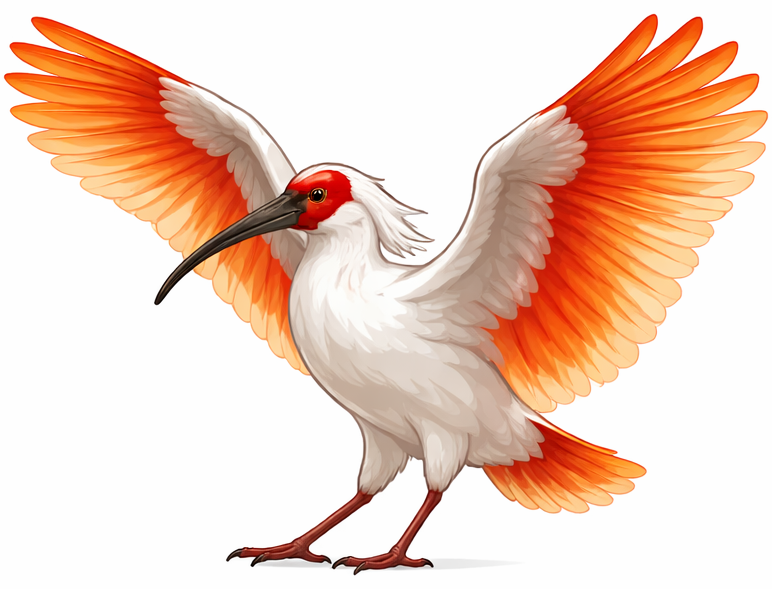}}\xspace}

\begin{document}

\twocolumn[
  \icmltitle{\icon Thoth: Mid-Training Bridges LLMs to Time Series Understanding}

  \icmlsetsymbol{equal}{*}

  \begin{icmlauthorlist}
    \icmlauthor{Jiafeng Lin}{equal,software}
    \icmlauthor{Yuxuan Wang}{equal,software}
    \icmlauthor{Jialong Wu}{equal,software}
    \icmlauthor{Huakun Luo}{software}
    \icmlauthor{Zhongyi Pei}{software}
    \icmlauthor{Jianmin Wang}{software}
  \end{icmlauthorlist}

  \icmlaffiliation{software}{
    School of Software, BNRist, Tsinghua University. Jiafeng Lin $<$lin-jf21@mails.tsinghua.edu.cn$>$. 
    Yuxuan Wang $<$wangyuxu22@mails.tsinghua.edu.cn$>$, Jialong Wu$<$wujialong0229@gmail.com$>$}

  \icmlcorrespondingauthor{Zhongyi Pei}{peizhyi@tsinghua.edu.cn}

  \icmlkeywords{large language models, time series, mid-training}

  \vskip 0.3in
]



\printAffiliationsAndNotice{}  

\begin{abstract}
Large Language Models (LLMs) have demonstrated remarkable success in general-purpose reasoning. However, they still struggle to understand and reason about time series data, which limits their effectiveness in decision-making scenarios that depend on temporal dynamics. In this paper, we propose \textbf{Thoth}, the first family of mid-trained LLMs with general-purpose time series understanding capabilities. As a pivotal intermediate stage, mid-training achieves task- and domain-agnostic alignment between time series and natural language, for which we construct \emph{Book-of-Thoth}, a high-quality, time-series-centric mid-training corpus. \emph{Book-of-Thoth} enables both time-series-to-text and text-to-time-series generation, equipping LLMs with a foundational grasp of temporal patterns. To better evaluate advanced reasoning capabilities, we further present KnoTS, a novel benchmark of knowledge-intensive time series understanding, designed for joint reasoning over temporal patterns and domain knowledge. Extensive experiments demonstrate that mid-training with \emph{Book-of-Thoth} enables Thoth to significantly outperform its base model and advanced LLMs across a range of time series question answering benchmarks. Moreover, Thoth exhibits superior capabilities when fine-tuned under data scarcity, underscoring the effectiveness of mid-training for time series understanding. Code is available at: \url{https://github.com/thuml/Thoth}.
\end{abstract}

\section{Introduction}
Large Language Models (LLMs) \cite{achiam2023gpt, zhao2023survey} have demonstrated remarkable capabilities in sequence modeling, complex reasoning \cite{wei2022chain}, and mathematical problem-solving \cite{azerbayev2023llemma}, largely driven by large-scale pre-training on massive textual corpora. However, real-world data are not exclusively represented in language but are instead often recorded as time series. Such time series data underpin a wide range of practical applications, including finance \cite{tsay2005analysis}, healthcare \cite{kaushik2020ai}, and transportation \cite{lippi2013short}. Bridging LLMs with time series data is therefore a critical step toward enabling LLM-based systems to reason over temporally grounded information and support more reliable decision-making in real-world scenarios.

Despite their promise, LLMs are fundamentally built upon language modeling and are not explicitly designed to model time series data, making it challenging for them to capture fine-grained temporal dependencies and the complex dynamics inherent in time series data. To address this gap, recent studies have predominantly relied on supervised fine-tuning (SFT), adapting LLMs to time series tasks using task-specific labeled datasets \cite{wang2025chattime, kong2025time}. While effective within task-specific scenarios, such approaches often require substantial labeled data and exhibit limited generalization across diverse temporal patterns, tasks, and domains, thereby constraining their scalability and highlighting the limitations of post-training.

\begin{figure*}[t]
  \vskip 0.2in
  \begin{center}
    \centerline{\includegraphics[width=\linewidth]{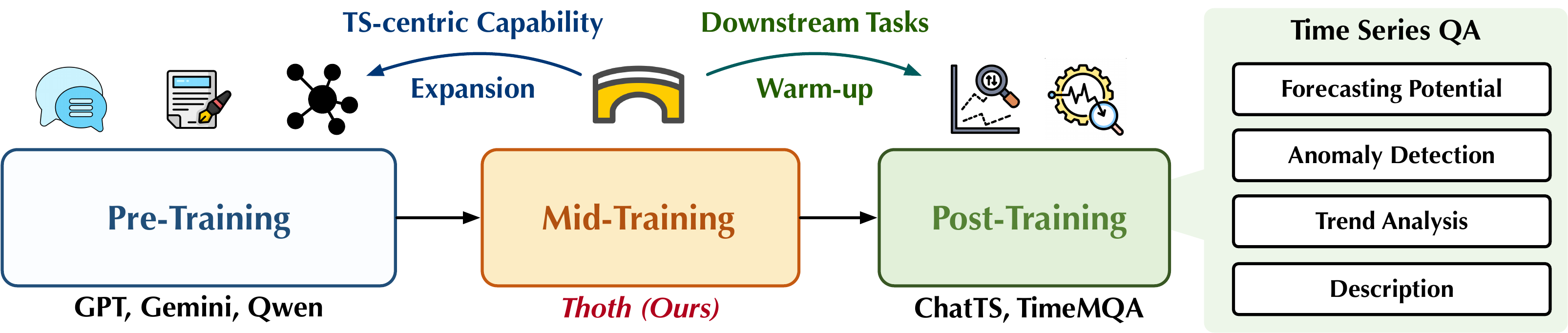}}
    \caption{Pivotal role of mid-training in bridging LLMs to time series understanding. Mid-training expands the capabilities of pre-trained LLMs while providing a critical transition and warm-up for subsequent post-training on specific time series question answering tasks.}
    \label{fig:intro}
  \end{center}
  \vspace{-10pt}
\end{figure*}

Recent advances in large language model research have witnessed a pronounced shift in focus toward mid-training. Positioned between large-scale pre-training and task-specific post-training, mid-training serves as an intermediate stage for reconciling general-purpose competencies with domain-specific specialization. As illustrated in Figure~\ref{fig:intro}, it acts as a bridge that expands the model’s knowledge coverage by incorporating a large-scale, high-quality time series-centric corpus. By preserving a proportion of general-purpose data, this stage maintains foundational competencies of LLMs while simultaneously amplifying targeted capabilities. Such intermediate adaptation better prepares the model for subsequent post-training, enabling a more stable and effective transition toward specialized task alignment.

Despite its growing success in domain expansion, mid-training remains largely unexplored in time series analysis, primarily because it relies heavily on suitable large-scale training data. In contrast to task-specific fine-tuning, mid-training aims to endow models with general-purpose capabilities, which requires training data to be broad in scope, diverse in structure, and grounded in language. However, most real-world time series multimodal datasets are narrowly focused on individual tasks, such as forecasting or classification, and are curated to directly map time series to downstream task labels, making them ill-suited for endowing LLMs with general time series understanding.

Based on these motivations, we introduce \textbf{Thoth}
\footnote{Thoth is the ancient Egyptian god of wisdom and the moon, traditionally depicted with the distinctive head of an ibis. He was credited with creating writing, reckoning time and seasons, and inventing the calendar. Many ancient Egyptian texts were attributed to him, collectively known as the Book of Thoth.}
, the first model family that leverages mid-training as an intermediate stage to bridge LLMs and time series understanding. To support task- and domain-agnostic alignment between temporal data and natural language, we construct \emph{Book-of-Thoth}, a large-scale, time-series-centric mid-training corpus. \emph{Book-of-Thoth} comprises complementary time-series-to-text and text-to-time-series generation as core mid-training tasks, exposing models to diverse temporal dynamics and their linguistic descriptions. Mid-training on \emph{Book-of-Thoth} enables Thoth to acquire transferable temporal representations while preserving the general reasoning capabilities of the base LLMs. To better evaluate the advanced reasoning capabilities of LLMs over both time series data and their inherent domain-specific knowledge, we introduce KnoTS, a novel benchmark of knowledge-intensive time series understanding. Experimentally, Thoth achieves substantial improvements on KnoTS and a wide range of time series question answering benchmarks, demonstrating strong few-shot and fine-tuned capabilities. These results indicate that Thoth serves as a critical intermediate step between general-purpose LLMs and advanced time series understanding.

Our contributions are summarized as follows:

\begin{itemize}
    \item We propose mid-training as a critical stage to endow LLMs with time series understanding.
    
    \item We automatically construct \emph{Book-of-Thoth}, a large-scale, time-series-centric corpus for aligning temporal data with natural language. It features time-series-to-text and text-to-time-series generation as core tasks.
    
    \item  We present Thoth, the first family of mid-trained LLMs with general-purpose time series understanding, which exhibits strong generalization across diverse time series question answering tasks.

    \item We introduce KnoTS, a novel benchmark for knowledge-intensive time series understanding.
\end{itemize}

\section{Related Work}

\subsection{Mid-Training on LLMs}

Mid-training has emerged as an important paradigm for LLMs. Positioned between large-scale pretraining and task-specific post-training, mid-training exposes models to high-quality specialized corpora, enabling domain expansion while preserving their general-purpose capabilities. 

The efficacy of mid-training is increasingly recognized across a wide range of domains. In the medical and legal domains, models such as PMC-LLaMA \cite{wu2024pmc} and SaulLM \cite{colombo2024saullm} illustrate how large-scale ingestion of domain-specific literature bridges the gap between general linguistic logic and highly structured professional expertise. In code intelligence, models like Codex \cite{chen2021evaluating}, and DeepSeek-Coder \cite{guo2024deepseek} leverage multi-stage training on heterogeneous programming corpora to enhance complex problem-solving abilities. Similarly, specialized datasets for the scientific sector, such as those used by GeoGalactica \cite{lin2023geogalactica}, improve technical reasoning in niche scientific fields.

Nevertheless, intensive training within these knowledge silos can sometimes lead to ``vertical forgetting'' \cite{tu2025survey}, where models may lose general capabilities. Prior work has explored regularization-based and rehearsal-based approaches to alleviate this issue.

\subsection{Time Series Language Models}

The intersection of time series modeling and LLMs has recently attracted increasing attention, particularly in the context of time series question answering (TSQA), driving the development of diverse datasets and specialized modeling techniques. Initially, \citet{merrill2024language} shows, via human-expert evaluations, that even highly capable LLMs exhibit limited time series understanding. ChatTS \cite{xie2024chatts} establishes a multimodal framework by treating time series as a native modality and leveraging synthetic data for alignment.  ChatTime \cite{wang2025chattime} innovatively models temporal data as a ``foreign language'', utilizing multi-stage training to enable zero-shot time series forecasting and bimodal interactions. To expand task diversity, Time-MQA \cite{kong2025time} introduces the TSQA dataset, which unifies multiple time series tasks into a comprehensive multi-task question-answering framework.  ITFormer \cite{wang2025itformer} further advances efficient integration by bridging temporal encoders with frozen LLMs through a lightweight instruction-tuning paradigm.  More recently, TIMEOMNI-1 \cite{guan2025timeomni} introduces TSR-SUITE to incorporate Chain-of-Thought (CoT) reasoning, shifting the focus from surface-level alignment to complex causal understanding and decision-making.

Despite the progress, most existing approaches rely heavily on instruction tuning on task-specific datasets. Recent advances in general-purpose LLMs suggest that strong domain performance often benefits from mid-training between large-scale pretraining and task-level adaptation \cite{tu2025survey}. However, such an intermediate stage for time series remains underexplored, which motivates the need for a mid-training phase tailored to time series understanding.

\begin{figure*}[t]
  \begin{center}
    \centerline{\includegraphics[width=\linewidth]{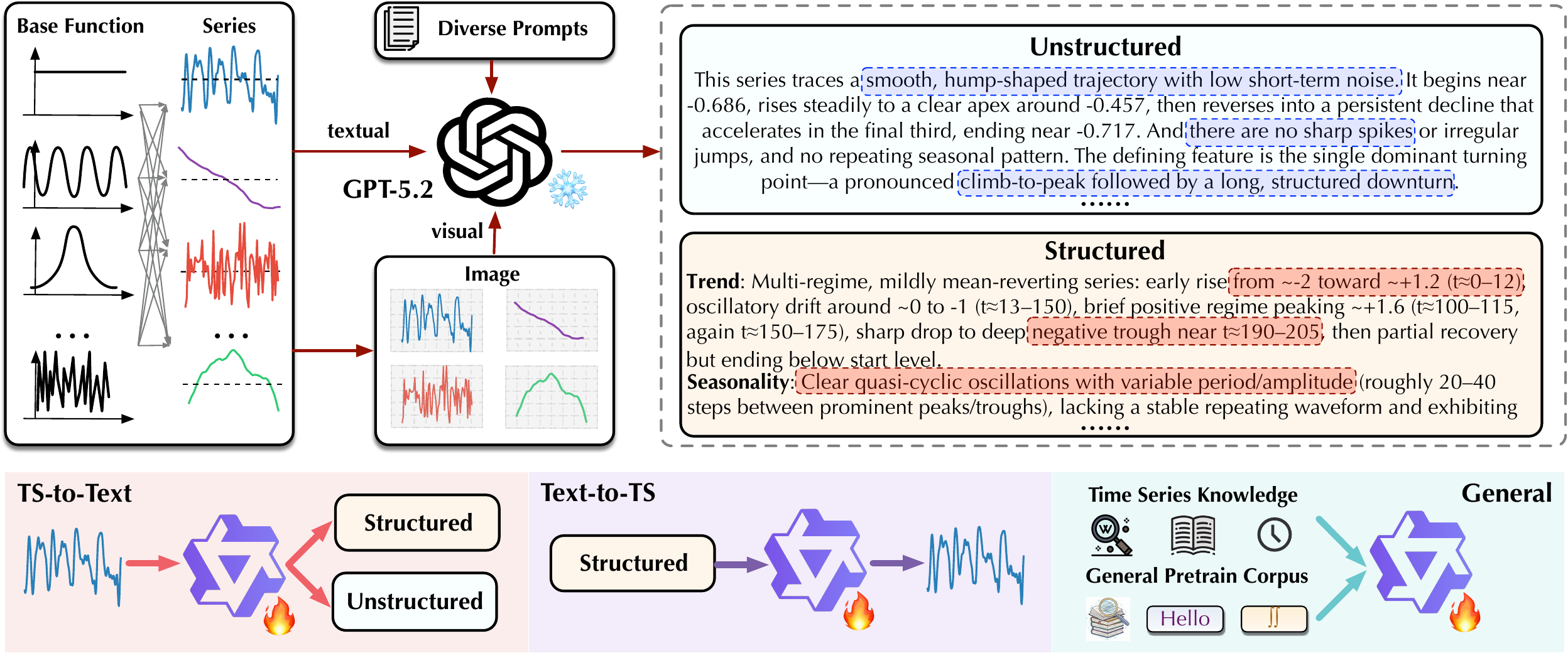}}
    \caption{The automated pipeline starts by synthesizing time series through combinations of diverse base functions. For each generated series, both the raw values and their visualizations are provided to GPT‑5.2 to produce a diverse range of time-series-to-text descriptions. These textual descriptions are then paired with the corresponding time series to construct complementary text-to-time-series generation data. In addition, a curated corpus of textual time series knowledge is incorporated to preserve the general-purpose capabilities of the pre-trained models during mid-training.}
    \label{fig:thoth}
  \end{center}
\vspace{-20pt}
\end{figure*}

\section{Problem Formulation}
 
Time series question answering is a prevalent paradigm for integrating LLMs with time series data. Given a univariate time series data $\mathbf{x}=\{\mathbf{x}_1, \mathbf{x}_2,\ldots,\mathbf{x}_T\}\in\mathbb{R}^T$ and a natural language query $\mathcal{Q}$, the goal is to generate a text-based response $\mathcal{A}$ that addresses $\mathcal{Q}$ by grounding the response in the evidence contained in $\mathbf{x}$. Formally, the task can be formulated as follows:
\begin{equation}
f_\theta(\mathcal{Q}, \mathbf{x})\rightarrow \mathcal{A},
\end{equation}
where $f_\theta$ denotes the model that models the query $\mathcal{Q}$ and the time series $\mathbf{x}$ to produce the corresponding answer $\mathcal{A}$.
Depending on the query, the response may consist of direct descriptions derived from observed values, or higher-level, context-aware reasoning that incorporates the semantic background provided in $\mathcal{Q}$ and justifies conclusions using evidence from $\mathbf{x}$. 

\section{Approach}

\subsection{Book of Thoth: Mid-Training Corpus}\label{sec:timecore}

We present \emph{Book-of-Thoth}, a comprehensive dataset designed for mid-training LLMs that integrates time series data with natural language descriptions. Compared to existing datasets, which are typically designed for narrowly defined objectives such as forecasting or pattern classification, \emph{Book-of-Thoth} is constructed to capture the bidirectional relationships between temporal patterns and textual representations, offering a robust foundation for time series understanding, as illustrated in Figure \ref{fig:thoth}.

Specifically, \emph{Book-of-Thoth} is composed of two interrelated components: (1) time series-to-text, in which each time series is annotated with a natural language caption highlighting salient temporal features, and (2) text-to-time series, in which textual descriptions are aligned with corresponding time series. To ensure scalability and consistency, an automated pipeline is developed to generate both types of high-quality time series–text pairs.

\vspace{-3pt}
\paragraph{Time Series-to-Text} The pipeline begins with time series generation. We employ KernelSynth \cite{ansari2024chronos}, a Gaussian-process–based method, to synthesize time series with rich and diverse temporal patterns. This \emph{domain-agnostic} synthetic generation provides a stable and controllable foundation for large-scale data construction. We then generate natural language descriptions from multiple perspectives through meticulously crafted design prompts with GPT-5.2. The resulting textual descriptions can be broadly categorized as structured or unstructured. Structured descriptions provide detailed, high-quality characterizations of time series properties from specific perspectives, such as trend, seasonality, and peak behavior. In contrast, unstructured descriptions consist of concise captions or summaries that emphasize the most salient temporal characteristics. These two forms of textual data are complementary, enabling the model to learn both fine-grained temporal patterns and high-level semantic abstractions.

\vspace{-3pt}
\paragraph{Text-to-Time Series} Based on the structured descriptions obtained in the time series-to-text stage, we construct text-to-time series pairs by reversing the time series and structured textual data. Each structured description specifies explicit temporal characteristics and is paired with its corresponding time series generated in the previous stage. This construction does not introduce additional synthesis and maintains a consistent alignment between the text and the time series.

\vspace{-3pt}
\paragraph{Textual Time Series Knowledge} Additionally, we incorporate open-source time series knowledge from two authoritative references: \emph{Forecasting: Principles and Practice} \cite{hyndman2018forecasting} and \emph{Time Series Analysis and Its Applications} \cite{shumway2006time}. These sources provide a solid theoretical foundation and practical guidelines for time series analysis and forecasting. We further employ GPT-4o-mini for data cleaning, including the removal of irrelevant content and logical refinement, to improve the quality and consistency of the training data.

\begin{figure}[h]
  \begin{center}
    \centerline{\includegraphics[width=1.0\columnwidth]{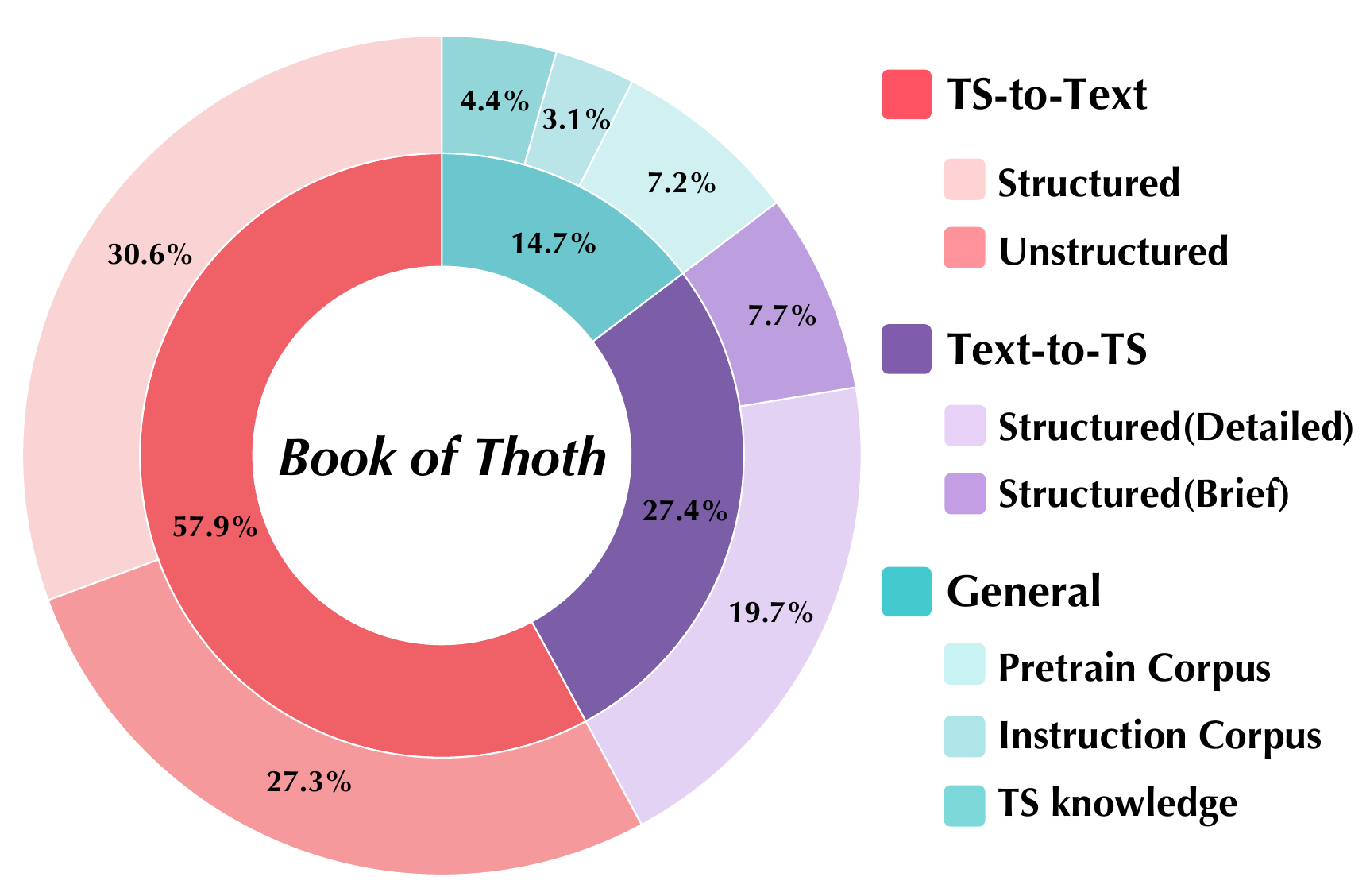}}
    \caption{
    Components of \emph{Book of Thoth} with 26.6M tokens.
    }
    \label{thoth-composition}
  \end{center}
  \vspace{-20pt}
\end{figure}

To mitigate catastrophic forgetting during mid-training, we adopt a standard mid-training strategy by mixing a small proportion of general-purpose LLM pre-training data into the training corpus. Specifically, we include samples from the C4 \cite{raffel2020exploring} and No Robots \cite{no_robots} datasets. This data mixture helps preserve general linguistic and world knowledge while enabling effective adaptation to time series–specific tasks. The proportions of each component are shown in Figure \ref{thoth-composition}.

\subsection{Thoth: Mid-Trained LLMs}

\paragraph{Architecture}

Thoth is a model family built upon the Qwen3 architecture and mid-trained on two model variants: Qwen3-30B-A3B-Instruct and Qwen3-8B \cite{yang2025qwen3}. Both variants share the core Qwen3 design, including Grouped Query Attention (GQA) and Rotary Positional Embeddings (RoPE) for efficient sequence modeling, as well as QK-Norm and RMSNorm for training stability.

\paragraph{Training}

We follow the default Qwen3 training recipe and keep the backbone architecture unchanged. All models are trained via full-parameter fine-tuning, with DeepSpeed ZeRO-3 employed for memory-efficient optimization. Thoth is optimized using AdamW with $(\beta_1, \beta_2) = (0.9, 0.999)$ under a cosine learning rate schedule. The peak learning rate is set to $1 \times 10^{-5}$, with a warmup ratio of $0.1$. Training is performed in \texttt{bfloat16} precision with a maximum sequence length of 4096 tokens. We use a global batch size of 32 and training runs for 3 epochs, resulting in approximately 1,600 optimization steps. Training is conducted on 8 NVIDIA A800 GPUs with 80GB of memory.

\section{KnoTS}

Existing time series question-answer (TSQA) benchmarks \cite{kong2025time, wang2025chattime} predominantly focus on narrow aspects of temporal understanding, such as trend analysis, outlier identification, short-term forecasting, and simple judgment-based queries. Notably, these benchmarks are often constructed either from a limited real-world corpus or through template-driven text paired with synthetic data. Consequently, such datasets exhibit low linguistic diversity and place minimal demands on general world knowledge, which, in turn, encourages models to rely on superficial mappings from recurring templates. This design further increases the risk of catastrophic forgetting when models are subsequently fine-tuned for specialized TSQA tasks.

\begin{figure}[h]
  \begin{center}
    \centerline{\includegraphics[width=1.0\columnwidth]{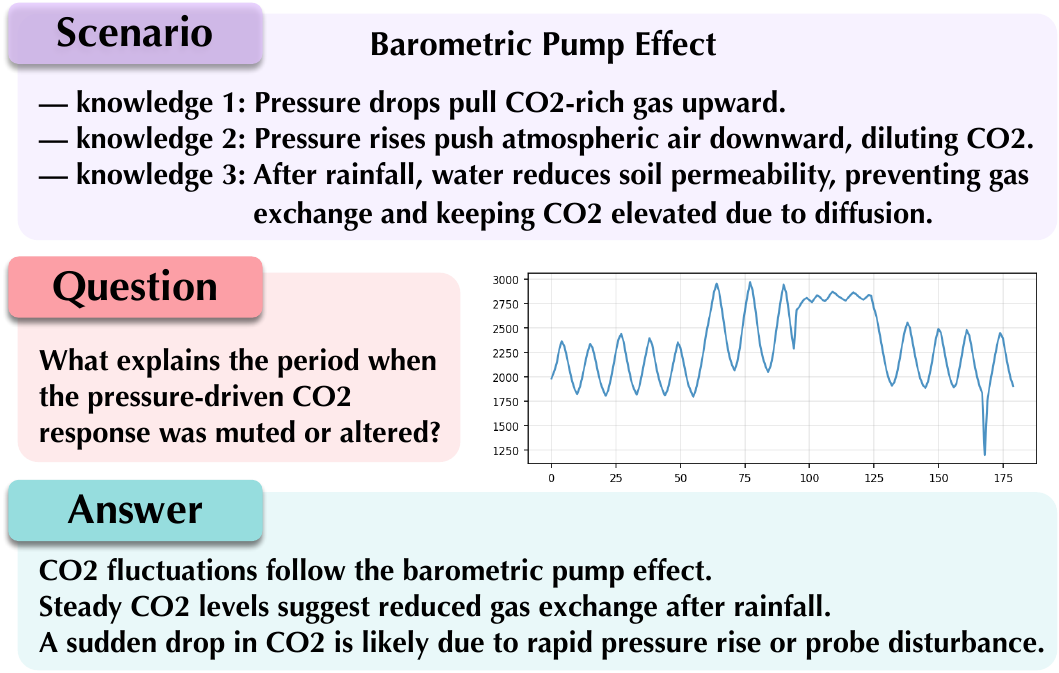}}
    \caption{
    A brief case of a reasoning problem from the KnoTS benchmark, the complete case can be found in the Appendix \ref{app:knots-cases}.
    }
    \label{KnoTS-brief-case}
  \end{center}
  \vspace{-20pt}
\end{figure}

To address these limitations, we introduce \textbf{KnoTS} (\textbf{Kno}wledge-intensive \textbf{T}ime \textbf{S}eries QA), a benchmark designed to evaluate joint reasoning over time series variations and domain knowledge, motivated by real-world scenarios in which interpreting time series data inherently depends on domain-specific context. We first identify a set of application domains in which time series data are fundamental to analysis and decision-making, including Finance, Healthcare, E-commerce, Environment, Social sectors. Within each domain, we prompt GPT to generate samples that consist of a time series paired with the corresponding domain-specific background information. For each sample, we then use GPT to generate a diverse set of question–answer pairs grounded in both the time series and the associated domain knowledge.
Subsequently, these candidates are reviewed and refined by human annotators to ensure factual accuracy, coherent alignment between the series and their explanations, and the sufficiency of the reasoning chain. The final benchmark consists of 300 carefully curated QA pairs. Figure \ref{KnoTS-brief-case} presents a simplified case in KnoTS. KnoTS comprises two complementary types of tasks: (1) reasoning tasks that focus on inferring temporal patterns informed by domain knowledge, and (2) decision-making tasks that involve predicting or making judgments by combining time series data with general domain knowledge.

\section{Experiment}

\paragraph{Datasets}

To comprehensively evaluate the time series understanding capabilities of Thoth, we adopt two widely recognized benchmarks: ChatTime \cite{wang2025chattime} and Time-MQA \cite{kong2025time}. ChatTime is designed for time series pattern recognition, focusing on four fundamental temporal characteristics: trend, volatility, seasonality, and outliers. All tasks are formulated as multiple-choice questions, with ground-truth answers precisely defined by the data generation process. Time-MQA is a real-world multi-task question-answering benchmark that encompasses seven diverse tasks, including Forecasting, Imputation, Classification, Anomaly Detection, Judgement, MCQ, and Open-Ended question answering. Together, these benchmarks provide a comprehensive evaluation framework, enabling us to assess both fine-grained discriminative understanding of temporal patterns and reasoning abilities in real-world time series scenarios.

\phantomsection
\begin{table*}[t]
\caption{Performance comparison across existing benchmarks. For the Time-MQA tasks, we report evaluation metrics including SMAPE for forecasting and imputation tasks, LLM-as-a-judge for open-ended tasks, and Accuracy (ACC) for others. For the ChatTime benchmark, the table reports the average results across all input lengths. The best results are highlighted in \boldres{bold} and the second best are \secondres{underlined}.}
\label{tab:main-results}
\centering
\resizebox{\textwidth}{!}{
\begin{tabular}{l|cccc | ccccccc}
\toprule
Model & \multicolumn{4}{c|}{ChatTime Benchmark} & \multicolumn{7}{c}{Time-MQA Benchmark} \\ \midrule
& Trend $\uparrow$ & Volatility $\uparrow$ & Seasonality $\uparrow$ & Outliers $\uparrow$ & Forecast $\downarrow$ & Impu. $\downarrow$ & Clas. $\uparrow$ & Anom. $\uparrow$ & Judge. $\uparrow$ & MCQ $\uparrow$ & Open. $\uparrow$\\
\midrule

\multicolumn{12}{c}{\cellcolor{myblue}\textit{Proprietary Models}} \\
\midrule
Gemini-3-Flash-Preview & 0.837 & \boldres{0.863} & 0.520 & \secondres{0.747} & 0.642 & 0.263 & 0.34 & 0.70 & 0.62 & 0.52 & 6.60 \\
GPT-4o-mini & 0.850 & 0.433 & 0.427 & 0.673 & 0.546 & 0.252 & 0.16 & \boldres{0.80} & \secondres{0.76} & 0.60 & 7.32 \\
Grok-4.1-Fast & 0.647 & 0.383 & \boldres{0.620} & \secondres{0.747} & 0.503 & 0.303 & 0.14 & 0.68 & \boldres{0.78} & 0.60 & \secondres{7.62} \\
\midrule
\multicolumn{12}{c}{\cellcolor{myred}\textit{Open-source Large Language Models}} \\
\midrule
Qwen3-235B-A22B-Instruct & 0.867 & 0.513 & 0.447 & 0.667 & \secondres{0.458} & \boldres{0.220} & 0.36 & \secondres{0.72} & 0.72 & \secondres{0.64} & \boldres{7.76} \\
Qwen3-30B-A3B-Instruct & 0.780 & 0.467 & 0.337 & 0.443 & 0.531 & 0.426 & 0.02 & 0.66 & 0.42 & 0.56 & 6.62 \\
Deepseek-R1-32B & 0.533 & 0.450 & 0.447 & 0.623 & 0.554 & 0.361 & 0.22 & 0.52 & 0.56 & 0.48 & 6.14 \\
Mistral-Small-24B-Instruct & 0.850 & 0.513 & 0.357 & 0.627 & 0.593 & 0.276 & 0.18 & 0.36 & 0.42 & 0.46 & 6.54 \\
Llama-3.1-8B-Instruct & 0.260 & 0.350 & 0.317 & 0.383 & 0.587 & 0.356 & \boldres{0.66} & 0.60 & 0.64 & 0.62 & 4.22 \\
Qwen3-8B & 0.163 & 0.220 & 0.100 & 0.447 & 0.671 & 0.533 & 0.02 & 0.48 & 0.62 & 0.54 & 6.28 \\
\midrule
\multicolumn{12}{c}{\cellcolor{mypurple}\textit{Open-source Vision Language Models}} \\
\midrule
Qwen3-VL-30B-A3B-Instruct & 0.860 & 0.460 & 0.350 & 0.470 & 0.610 & 0.278 & 0.12 & 0.62 & 0.72 & 0.48 & 7.08 \\
Mistral-Small-3.2-24B-Instruct & 0.430 & 0.373 & 0.597 & 0.563 & 0.578 & 0.270 & \secondres{0.38} & 0.66 & 0.74 & 0.52 & 6.54 \\
\midrule
\rowcolor{gray!20} \multicolumn{12}{c}{\textit{Task-Specific Time Series Language Models}} \\
\midrule
ChatTS-14B & - & - & - & - & 1.166 & - & 0.24 & 0.44 & 0.58 & 0.46 & 4.18 \\
ChatTime-7B-Chat & \textcolor{gray!80}{0.980} & \textcolor{gray!80}{0.870} & \textcolor{gray!80}{0.730} & \textcolor{gray!80}{0.963} & - & - & - & - & - & - & - \\
Time-MQA(LLaMA3-8B) & 0.653 & 0.433 & 0.300 & 0.393 & \textcolor{gray!80}{0.528} & \textcolor{gray!80}{0.307} & \textcolor{gray!80}{0.44} & \textcolor{gray!80}{0.58} & \textcolor{gray!80}{0.72} & \textcolor{gray!80}{0.64} & \textcolor{gray!80}{4.80} \\
Time-MQA(Qwen2.5-7B) & 0.297 & 0.370 & 0.343 & 0.390 & \textcolor{gray!80}{0.480} & \textcolor{gray!80}{0.251} & \textcolor{gray!80}{0.28} & \textcolor{gray!80}{0.64} & \textcolor{gray!80}{0.66} & \textcolor{gray!80}{0.46} & \textcolor{gray!80}{5.06} \\
\midrule
\multicolumn{12}{c}{\cellcolor{mygreen}\textit{Ours}} \\
\midrule
\textbf{Thoth-8B} & \boldres{0.973} & 0.750 & \secondres{0.603} & \boldres{0.753} & 0.522 & 0.333 & 0.10 & 0.58 & 0.72 & \secondres{0.64} & 5.98 \\
\textbf{Thoth-30B-A3B} & \secondres{0.957} & \secondres{0.770} & 0.573 & 0.740 & \boldres{0.432} & \secondres{0.247} & 0.02 & 0.70 & \boldres{0.78} & \boldres{0.70} & 7.42 \\
\bottomrule
\end{tabular}
}
\vspace{-5pt}
\end{table*}

\paragraph{Baselines}

We evaluate Thoth against 15 advanced baseline models, spanning four categories: (1) \emph{Proprietary models}: Gemini-3-Flash-Preview, GPT-4o-mini, Grok-4.1-Fast, (2) \emph{Open-source Large Language Models}: Qwen3-235B-A22B-Instruct \cite{yang2025qwen3}, Qwen3-30B-A3B-Instruct \cite{yang2025qwen3}, Deepseek-R1-32B \cite{guo2025deepseek}, Mistral-Small-24B-Instruct, LLaMA3.1-8B-Instruct, Qwen3-8B \cite{yang2025qwen3}, (3) \emph{Open-source Vision Language Models}: Qwen3-VL-30B-A3B-Instruct \cite{bai2025qwen3vltechnicalreport}, Mistral-Small-3.2-24B-Instruct, and (4) \emph{Task-Specific Time Series Language Models}: ChatTS-14B \cite{xie2024chatts}, Time-MQA (Llama3-8B) \cite{kong2025time}, Time-MQA (Qwen2.5-7B) \cite{kong2025time}, ChatTime-7B-Chat \cite{wang2025chattime}.

\paragraph{Evaluation Details}

To evaluate the time series understanding capabilities of Thoth while ensuring a fair comparison with baselines, we employ the widely recognized \emph{n-shot in-context learning} protocol \cite{brown2020language}. Specifically, we provide 3-5 fixed examples per task as demonstrations in the prompt for all models. This range is chosen to maintain an optimal balance between context window constraints and the need for class representation across different categories. These examples serve to guide the LLM in following complex instructions and adhering to specific response formats, such as structured multiple-choice answers.

In terms of evaluation metrics, the four sub-tasks within ChatTime are evaluated using Accuracy (ACC), as all tasks are formulated as multiple-choice questions. For Time-MQA, metrics are task-specific: Forecasting and Imputation are evaluated using Symmetric Mean Absolute Percentage Error (SMAPE), which is robust to scale variations commonly observed in real-world time series. Classification, Anomaly Detection, Judgement, and Multiple-Choice Question tasks are evaluated via Accuracy (ACC). For the Open-Ended task and our newly introduced KnoTS, where explicit ground-truth labels are unavailable, we employ the advanced Gemini-3-Pro-Preview as an automated evaluator, which assigns scores along multiple dimensions, including factual correctness, temporal logic, completeness, and conciseness.

\subsection{Main Results}

\begin{figure}[h]
  \begin{center}
    \centerline{\includegraphics[width=1.0\columnwidth]{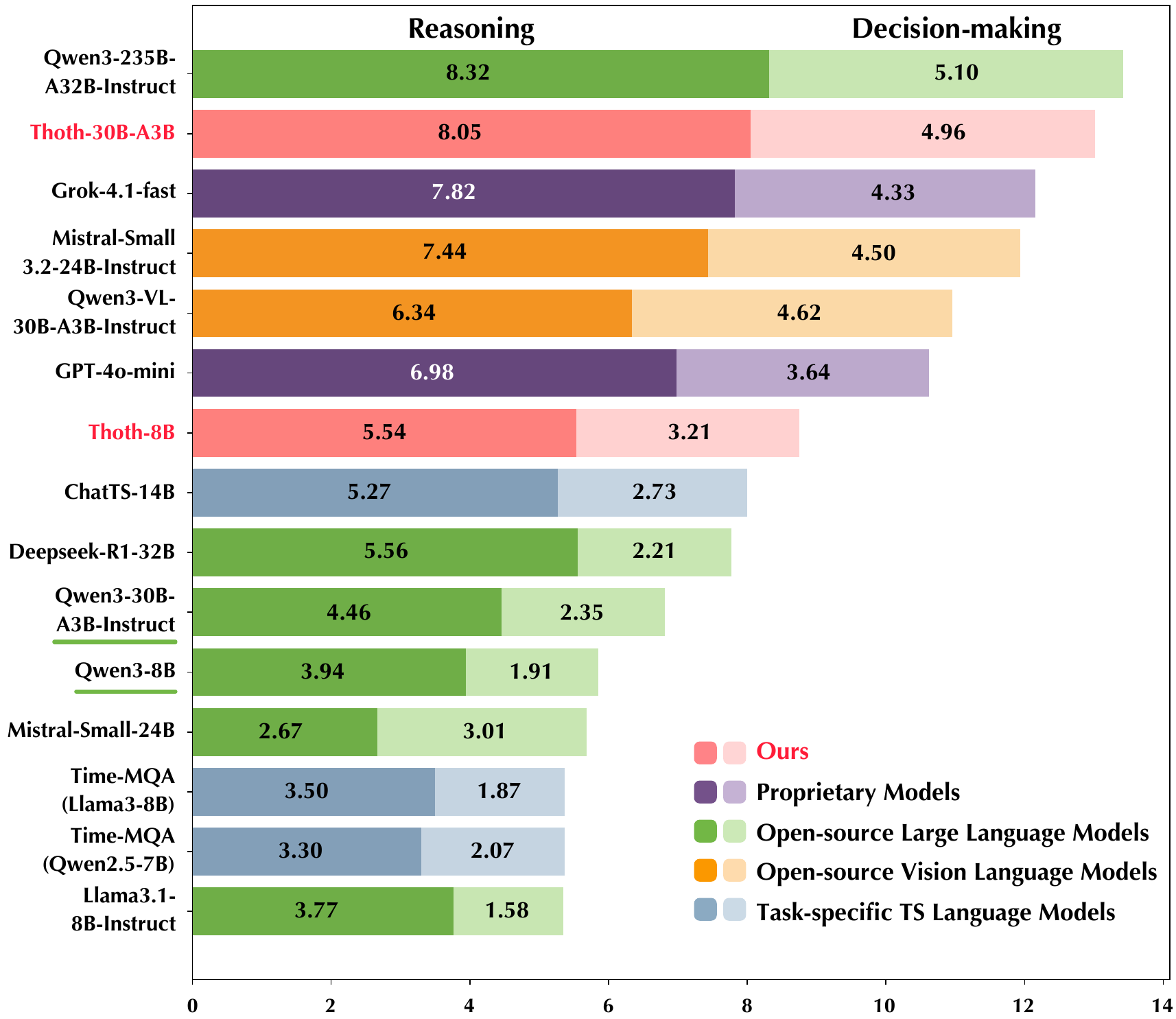}}
    \caption{
    Performance of various models in reasoning and decision-making on the KnoTS benchmark. All evaluations are conducted using Gemini-3-Pro-Preview with scores ranging from 0.0 to 10.0.
    }
    \label{knots-results}
  \end{center}
  \vspace{-23pt}
\end{figure}

\begin{figure*}[t]
  \begin{center}
    \centerline{\includegraphics[width=1.0\linewidth]{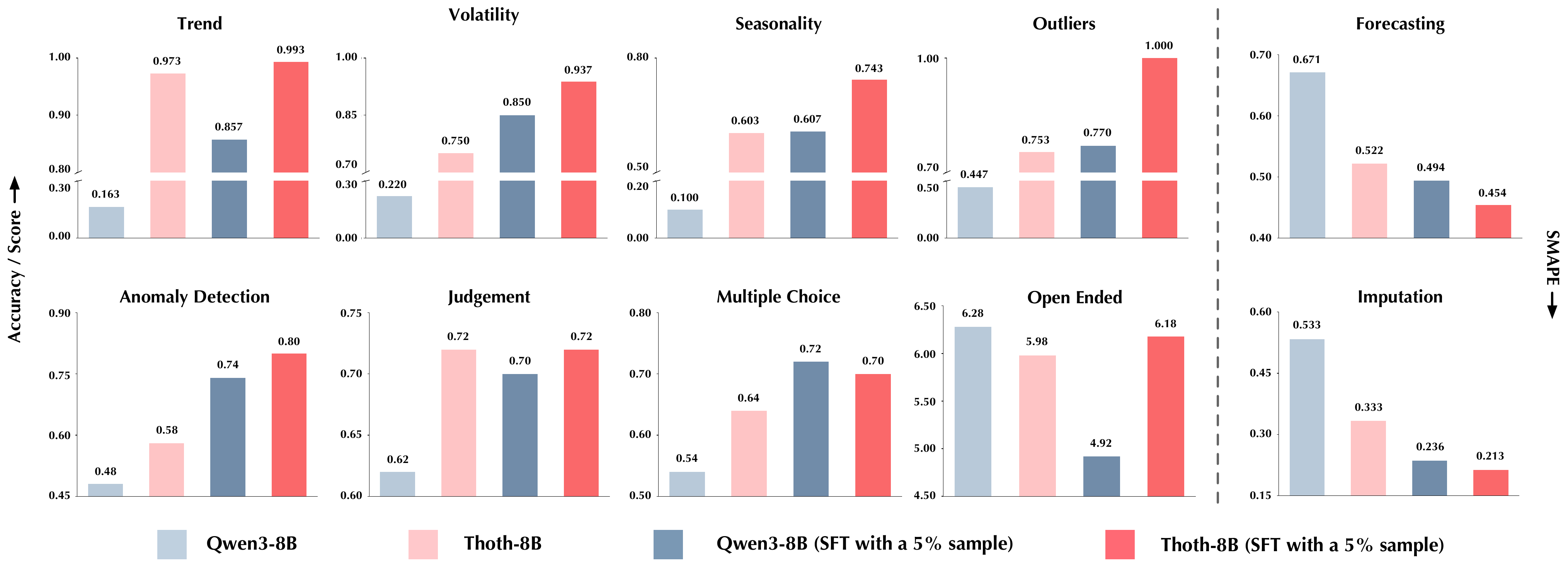}}
    \caption{
    Performance of Supervised Fine-tuning on ChatTime and Time-MQA benchmark using a 5\% sampled training set. For each benchmark, an equal proportion of data is randomly selected from each sub-task for training, with no separate model trained per sub-task.
    }
    \label{fig:results-fewshot}
  \end{center}
  \vspace{-20pt}
\end{figure*}

Experimental results are presented in Table \ref{tab:main-results}. Thoth demonstrates remarkable capabilities on time series question answering. Thoth-30B-A3B achieves overall performance comparable to, and in some cases exceeding, the 235B parameter Qwen3 model. Meanwhile, Thoth-8B performs on par with open-source models at around 30B parameters, and substantially outperforms the Qwen3-8B base model.

Downstream performance for Time Series Language Models is shown in gray. These models are directly supervised fine-tuned on downstream datasets, which often leads to pronounced cross-task imbalance and unstable performance across different tasks. Notably, Open-source VLMs that incorporate time series plots as visual inputs consistently outperform parameter-matched LLMs, highlighting the importance of visual information for time series understanding. This observation further underscores the effectiveness of constructing \emph{Book-of-Thoth} with dual inputs of time series text and images when using GPT-5.2 for data generation.

The ChatTime benchmark further stratifies evaluations by input length and the complete results are provided in Appendix \ref{app:chattime-full-results}. As LLM-based generation is inherently probabilistic, we additionally conduct an error bar analysis to quantify uncertainty, with detailed results reported in Appendix \ref{app:error-bar}.

Beyond these public time-series question-answering benchmarks, we further conduct experiments on our proposed KnoTS to assess the capabilities of joint reasoning over time series and domain knowledge and decision-making. The evaluation results are scored by Gemini-3-Pro-Preview, one of the most powerful LLMs currently available. As a result, Gemini-3-Flash-Preview is excluded from the baseline models, as models within the same series are likely to share substantial portions of their training corpus.

As shown in Figure \ref{knots-results}, Thoth-30B-A3B demonstrates superior performance, outperforming powerful LLMs such as Grok-4.1-Fast and VLMs, and is only slightly surpassed by the Qwen3 model, which has a total parameter size of 235B. Although Thoth-8B is constrained by the smaller parameter scale of its base model, which limits its overall reasoning capacity, it substantially enhances time series understanding through mid-training. As a result, Thoth-8B outperforms most open-source LLMs in the 30B parameter range, as well as all existing Time Series Language Models.

\subsection{Supervised Fine-tuning Results}

As a mid-trained LLM, Thoth is designed to provide a critical warm-up for post-training and specific downstream tasks. 
In this section, we fine-tune Thoth on a randomly sampled 5\% subset from Time-MQA and ChatTime respectively, while preserving the original data distribution within each benchmark. We compare the results with those obtained by directly fine-tuning the original Qwen3-8B model using the same data subset. As illustrated in Figure \ref{fig:results-fewshot}, after supervised fine-tuning, Thoth significantly outperforms the baseline LLMs on the vast majority of tasks, demonstrating the effectiveness of Thoth for warm-up downstream tasks.

An interesting observation is that performance on open-ended tasks decreases sharply after fine-tuning Qwen3-8B, suggesting that directly fine-tuning on multiple downstream tasks can induce an undesirable capability trade-off, where gains on certain skills come at the expense of others. In contrast, fine-tuned Thoth-8B achieves a stable performance improvement. This is attributed to its mid-training phase, which allows the model to leverage its established understanding to assimilate task-specific knowledge during multi-task fine-tuning rather than resorting to rote memorization.

\subsection{Data Scaling}

\begin{figure}[ht]
  \vspace{-5pt}
  \begin{center}
    \centerline{\includegraphics[width=0.95\columnwidth]{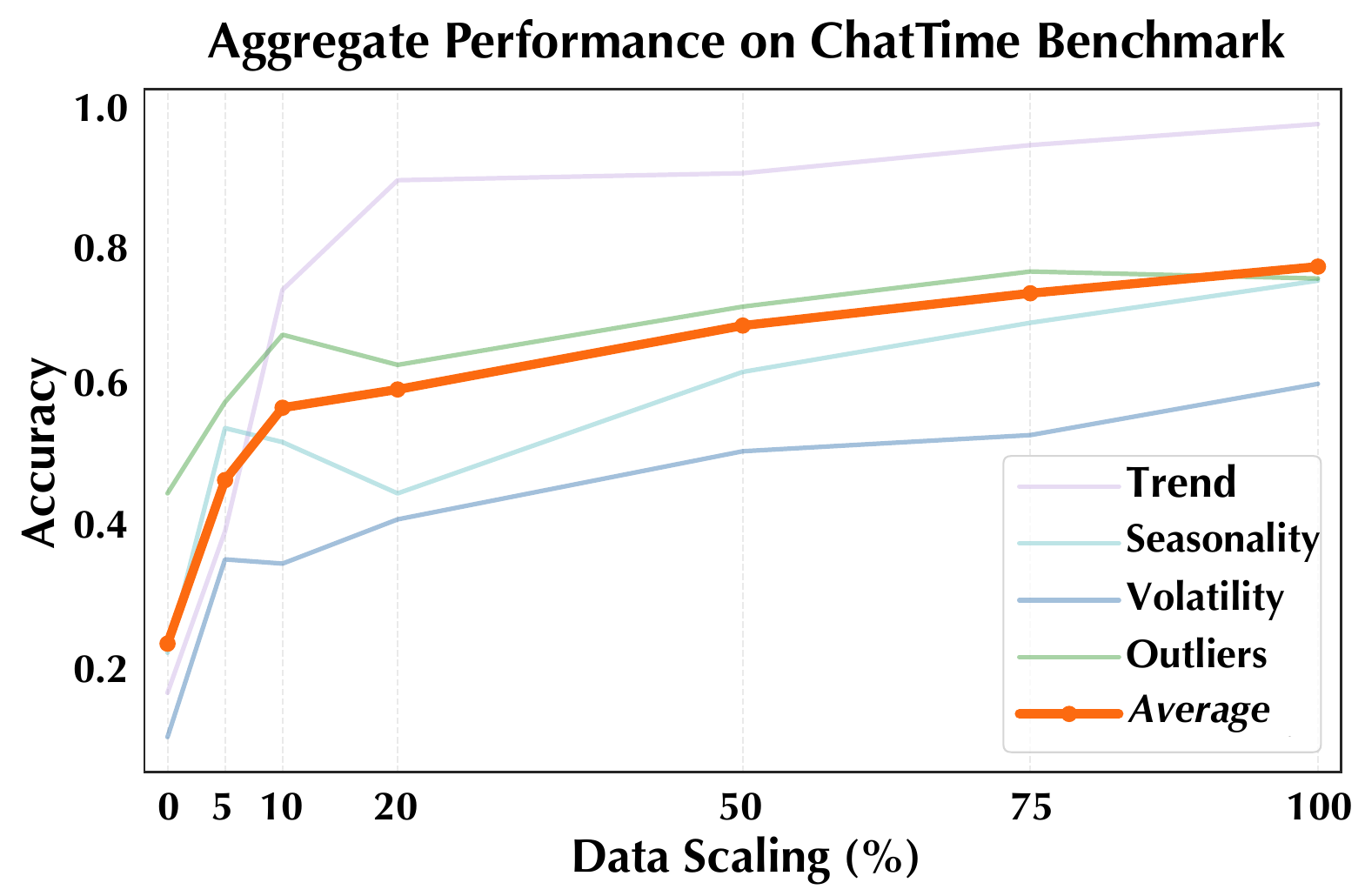}}
    \caption{
    Aggregate performance of Thoth-8B on ChatTime benchmark at different mid-training data scales. As the data scale increases, Thoth-8B exhibits stronger average performance.
    }
    \label{scaling-chattime}
  \end{center}
  \vspace{-20pt}
\end{figure}

In this section, we examine the empirical scaling behavior of our model by conducting mid-training experiments across varying data regimes. Specifically, we utilize subsets of \emph{Book-of-Thoth} representing 10\%, 20\%, 50\%, 75\%, and 100\% of the total volume, while maintaining a consistent data distribution. The results presented in Figure \ref{scaling-chattime}, reveal a clear trend that the model performance improves steadily as the dataset size increases, highlighting the significant potential of data scaling.

\begin{table*}[t]
\caption{Ablation results on \emph{Book-of-Thoth}. We separately remove the three types of corpora, \emph{Time series-to-text pairs} (TS-to-Text), \emph{Text-to-time series pairs} (Text-to-TS), and \emph{General pre-training data} (General), for training and compare them with Thoth-8b.}\label{tab:ablation}
\centering
\resizebox{\textwidth}{!}{
\begin{tabular}{l|cccc | ccccccc | cc}
\toprule
Model & \multicolumn{4}{c|}{ChatTime Benchmark} & \multicolumn{7}{c|}{Time-MQA Benchmark} & \multicolumn{2}{c}{KnoTS Benchmark} \\ \midrule
& Trend $\uparrow$ & Volatility $\uparrow$ & Seasonality $\uparrow$ & Outliers $\uparrow$ & Forecast $\downarrow$ & Impu. $\downarrow$ & Clas. $\uparrow$ & Anom. $\uparrow$ & Judge. $\uparrow$ & MCQ $\uparrow$ & Open. $\uparrow$ & Reason $\uparrow$ & Decision $\uparrow$\\
\midrule
Qwen3-8B & 0.163 & 0.220 & 0.100 & 0.447 & 0.671 & 0.533 & 0.02 & 0.48 & 0.62 & 0.54 & 6.28 & 3.94 & 1.91 \\
\midrule
-- TS-to-Text & 0.360 & 0.487 & 0.347 & 0.453 & \textbf{0.495} & \textbf{0.286} & 0.14 & 0.48 & 0.68 & 0.46 & 3.32 & 3.13 & 1.45 \\
-- Text-to-TS & 0.967 & 0.677 & \textbf{0.630} & 0.733 & 0.532 & 0.476 & 0.12 & 0.56 & 0.68 & 0.58 & 5.10 & 5.23 & 3.04 \\
-- General & 0.963 & 0.547 & 0.387 & \textbf{0.763} & 0.538 & 0.363 & \textbf{0.16} & \textbf{0.58} & 0.70 & 0.60 & \textbf{6.12} & 5.30 & 3.06 \\
\midrule
Thoth-8B & \textbf{0.973} & \textbf{0.750} & 0.603 & 0.753 & 0.522 & 0.333 & 0.100 & \textbf{0.58} & \textbf{0.72} & \textbf{0.64} & 5.98 & \textbf{5.54} & \textbf{3.21} \\
\bottomrule
\end{tabular}
}
\end{table*}

\subsection{Ablation Study}\label{sec:ablation}

To verify the effectiveness of \emph{Book-of-Thoth} components, we provide detailed
ablations by removing each of the three components individually. The results are listed in Table \ref{tab:ablation}. Upon removing the \emph{TS-to-Text} component, since the remaining dataset predominantly consists of time series generation, the performance of two tasks related to time series generation, forecasting and imputation, improves, although their overall performance remains suboptimal.

Interestingly, while \emph{Text-to-TS} enhances overall time series understanding, the improvement is less pronounced compared to \emph{TS-to-Text}. This discrepancy likely arises from the fact that reconstructing precise time series from text presents a much greater challenge than generating a textual description from time series. In contexts with limited model capacity, simpler auxiliary tasks often yield greater benefits. A typical instance is the Masked Language Modeling (MLM) task in BERT \cite{devlin2019bert}, which, by capturing bidirectional contextual information at the token level, proves highly efficient for feature extraction in smaller models. However, as model size increased, the more complex Next Token Prediction (NTP) task, following the GPT paradigm, demonstrates its advantages, aligning with the enhanced generative capabilities of larger models \cite{brown2020language}. Although the improvement from \emph{Text-to-TS} currently may not be as pronounced as that from \emph{TS-to-Text}, it still offers valuable enhancement and remains an indispensable potential for advancing time series understanding.

\subsection{Case study}

\begin{figure}[ht]
  \begin{center}
    \centerline{\includegraphics[width=1\columnwidth]{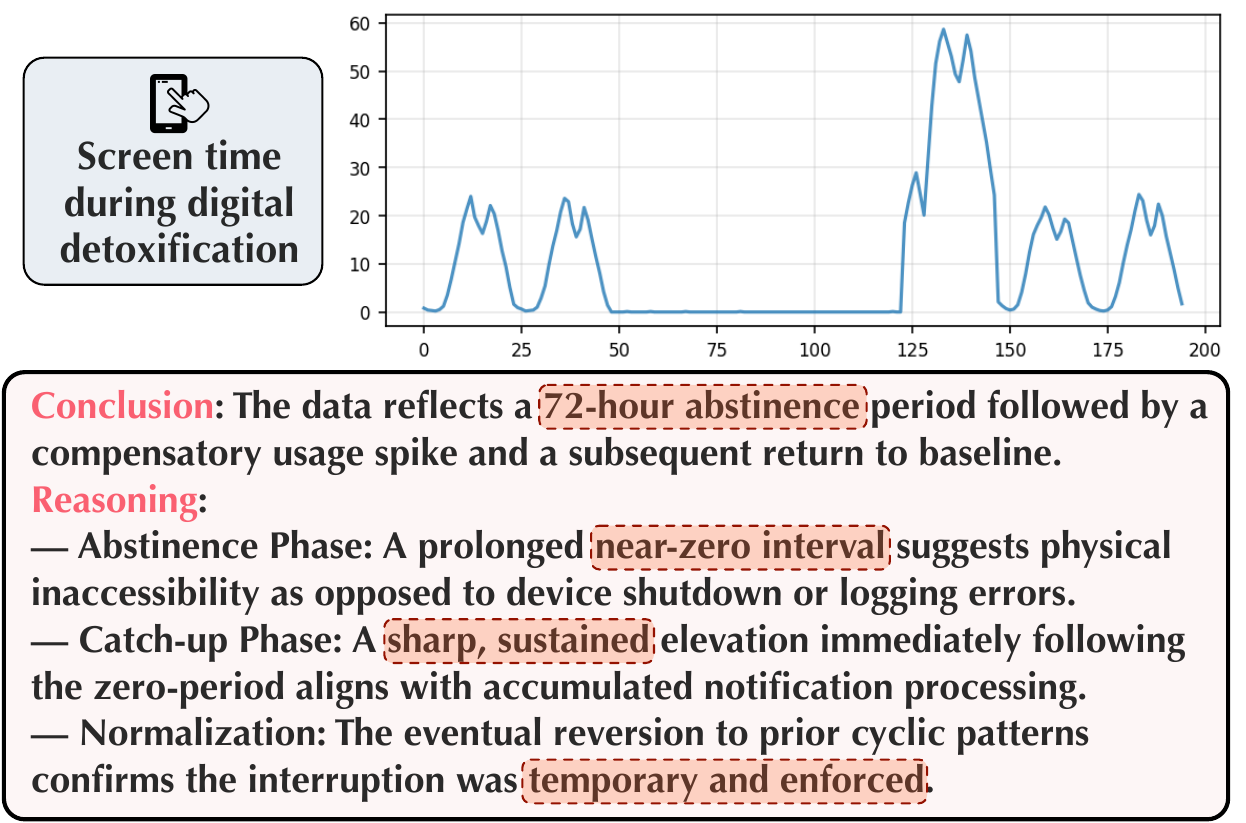}}
    \vspace{3pt}
    \caption{
    Case study of Thoth-30B-A3B model inference on the KnoTS benchmark. Detailed comparative results of various models can be found in Appendix \ref{app:showcases} for further reference.
    }
    \label{showcase-main}
  \end{center}
  \vspace{-23pt}
\end{figure}

This section provides a case study in Figure \ref{showcase-main}, showing the simplified inference results of Thoth-30B-A3B on the KnoTS benchmark. As illustrated, the model infers the user’s specific behaviors across three distinct stages based on time series analysis. These results are consistent with the ground truth, demonstrating Thoth’s robust time series reasoning capabilities in domain-specific scenarios.

\section{Discussions}

Given the prevalence of time series data in real-world applications, empowering LLMs with time series understanding capability is of increasing importance. This paper treats mid-training as a practical bridge that adapts LLM toward time series understanding. To this end, we construct \emph{Book-of-Thoth}, a large-scale, task- and domain-agnostic, time series–centric corpus that aligns time series with natural language, and use it to train Thoth, a family of general-purpose LLMs with substantially improved time series understanding. To better evaluate the time series reasoning capabilities of LLMs, we further propose a knowledge-intensive time series QA benchmark that emphasizes reasoning and decision-making grounded in temporal evidence. Extensive experiments demonstrate that Thoth delivers strong few-shot performance on a broad range of time series tasks. Moreover, a supervised fine-tuning study shows that mid-training effectively warms up downstream time-series QA, enabling consistent gains with limited task-specific data. Analysis experiment demonstrates strong scalability and provides valuable insights.

\vspace{-10pt}

\paragraph{Limitations and Future Work}

This paper focuses on bridging LLMs to time series understanding via mid-training. Looking ahead, it is valuable to explore further advancements in post-training to enhance time series understanding and reasoning. Figure 1 illustrates the powerful capability of mid-training in warming up downstream tasks. However, it is evident that direct fine-tuning of multi-task learning for downstream tasks may lead to performance degradation in certain tasks, due to the imbalanced requirements across tasks. Therefore, reinforcement learning, which has been widely proven effective in prior work on question-answering tasks, is worth investigating.

Table \ref{tab:ablation} demonstrates the impact of different components of the \emph{Book-of-Thoth} framework on model performance. The text-to-time series training task has shown promising potential; however, as discussed in Section \ref{sec:ablation}, there is still room for improvement on larger-scale models and datasets. Moreover, it remains an open question whether the training signal of generated time series should be based on its alignment with textual information, rather than the traditional next-token prediction loss.

\nocite{langley00}

\bibliography{example_paper}

@inproceedings{langley00,
 author    = {P. Langley},
 title     = {Crafting Papers on Machine Learning},
 year      = {2000},
 pages     = {1207--1216},
 editor    = {Pat Langley},
 booktitle     = {Proceedings of the 17th International Conference
              on Machine Learning (ICML 2000)},
 address   = {Stanford, CA},
 publisher = {Morgan Kaufmann}
}

@article{wu2024pmc,
  title={PMC-LLaMA: toward building open-source language models for medicine},
  author={Wu, Chaoyi and Lin, Weixiong and Zhang, Xiaoman and Zhang, Ya and Xie, Weidi and Wang, Yanfeng},
  journal={Journal of the American Medical Informatics Association},
  volume={31},
  number={9},
  pages={1833--1843},
  year={2024},
  publisher={Oxford Academic}
}

@article{colombo2024saullm,
  title={Saullm-7b: A pioneering large language model for law},
  author={Colombo, Pierre and Pires, Telmo Pessoa and Boudiaf, Malik and Culver, Dominic and Melo, Rui and Corro, Caio and Martins, Andre FT and Esposito, Fabrizio and Raposo, Vera L{\'u}cia and Morgado, Sofia and others},
  journal={arXiv preprint arXiv:2403.03883},
  year={2024}
}

@article{chen2021evaluating,
  title={Evaluating large language models trained on code},
  author={Chen, Mark},
  journal={arXiv preprint arXiv:2107.03374},
  year={2021}
}

@article{guo2024deepseek,
  title={DeepSeek-Coder: When the Large Language Model Meets Programming--The Rise of Code Intelligence},
  author={Guo, Daya and Zhu, Qihao and Yang, Dejian and Xie, Zhenda and Dong, Kai and Zhang, Wentao and Chen, Guanting and Bi, Xiao and Wu, Yu and Li, YK and others},
  journal={arXiv preprint arXiv:2401.14196},
  year={2024}
}

@article{lin2023geogalactica,
  title={Geogalactica: A scientific large language model in geoscience},
  author={Lin, Zhouhan and Deng, Cheng and Zhou, Le and Zhang, Tianhang and Xu, Yi and Xu, Yutong and He, Zhongmou and Shi, Yuanyuan and Dai, Beiya and Song, Yunchong and others},
  journal={arXiv preprint arXiv:2401.00434},
  year={2023}
}

@article{azerbayev2023llemma,
  title={Llemma: An open language model for mathematics},
  author={Azerbayev, Zhangir and Schoelkopf, Hailey and Paster, Keiran and Santos, Marco Dos and McAleer, Stephen and Jiang, Albert Q and Deng, Jia and Biderman, Stella and Welleck, Sean},
  journal={arXiv preprint arXiv:2310.10631},
  year={2023}
}

@article{xie2024chatts,
  title={Chatts: Aligning time series with llms via synthetic data for enhanced understanding and reasoning},
  author={Xie, Zhe and Li, Zeyan and He, Xiao and Xu, Longlong and Wen, Xidao and Zhang, Tieying and Chen, Jianjun and Shi, Rui and Pei, Dan},
  journal={arXiv preprint arXiv:2412.03104},
  year={2024}
}

@inproceedings{wang2025chattime,
  title={Chattime: A unified multimodal time series foundation model bridging numerical and textual data},
  author={Wang, Chengsen and Qi, Qi and Wang, Jingyu and Sun, Haifeng and Zhuang, Zirui and Wu, Jinming and Zhang, Lei and Liao, Jianxin},
  booktitle={Proceedings of the AAAI Conference on Artificial Intelligence},
  volume={39},
  number={12},
  pages={12694--12702},
  year={2025}
}

@article{wang2025itformer,
  title={ITFormer: Bridging Time Series and Natural Language for Multi-Modal QA with Large-Scale Multitask Dataset},
  author={Wang, Yilin and Lei, Peixuan and Song, Jie and Hao, Yuzhe and Chen, Tao and Zhang, Yuxuan and Jia, Lei and Li, Yuanxiang and Wei, Zhongyu},
  journal={arXiv preprint arXiv:2506.20093},
  year={2025}
}

@article{kong2025time,
  title={Time-mqa: Time series multi-task question answering with context enhancement},
  author={Kong, Yaxuan and Yang, Yiyuan and Hwang, Yoontae and Du, Wenjie and Zohren, Stefan and Wang, Zhangyang and Jin, Ming and Wen, Qingsong},
  journal={arXiv preprint arXiv:2503.01875},
  year={2025}
}

@article{guan2025timeomni,
  title={TimeOmni-1: Incentivizing Complex Reasoning with Time Series in Large Language Models},
  author={Guan, Tong and Meng, Zijie and Li, Dianqi and Wang, Shiyu and Yang, Chao-Han Huck and Wen, Qingsong and Liu, Zuozhu and Siniscalchi, Sabato Marco and Jin, Ming and Pan, Shirui},
  journal={arXiv preprint arXiv:2509.24803},
  year={2025}
}

@article{chen2025mtbench,
  title={Mtbench: A multimodal time series benchmark for temporal reasoning and question answering},
  author={Chen, Jialin and Feng, Aosong and Zhao, Ziyu and Garza, Juan and Nurbek, Gaukhar and Qin, Cheng and Maatouk, Ali and Tassiulas, Leandros and Gao, Yifeng and Ying, Rex},
  journal={arXiv preprint arXiv:2503.16858},
  year={2025}
}

@article{zhang2025insight,
  title={Insight miner: A time series analysis dataset for cross-domain alignment with natural language},
  author={Zhang, Yunkai and Zhang, Yawen and Zheng, Ming and Chen, Kezhen and Gao, Chongyang and Ge, Ruian and Teng, Siyuan and Jelloul, Amine and Rao, Jinmeng and Guo, Xiaoyuan and others},
  journal={arXiv preprint arXiv:2512.11251},
  year={2025}
}

@article{achiam2023gpt,
  title={Gpt-4 technical report},
  author={Achiam, Josh and Adler, Steven and Agarwal, Sandhini and Ahmad, Lama and Akkaya, Ilge and Aleman, Florencia Leoni and Almeida, Diogo and Altenschmidt, Janko and Altman, Sam and Anadkat, Shyamal and others},
  journal={arXiv preprint arXiv:2303.08774},
  year={2023}
}

@article{zhao2023survey,
  title={A survey of large language models},
  author={Zhao, Wayne Xin and Zhou, Kun and Li, Junyi and Tang, Tianyi and Wang, Xiaolei and Hou, Yupeng and Min, Yingqian and Zhang, Beichen and Zhang, Junjie and Dong, Zican and others},
  journal={arXiv preprint arXiv:2303.18223},
  volume={1},
  number={2},
  year={2023}
}

@book{tsay2005analysis,
  title={Analysis of financial time series},
  author={Tsay, Ruey S},
  year={2005},
  publisher={John wiley \& sons}
}

@article{kaushik2020ai,
  title={AI in healthcare: time-series forecasting using statistical, neural, and ensemble architectures},
  author={Kaushik, Shruti and Choudhury, Abhinav and Sheron, Pankaj Kumar and Dasgupta, Nataraj and Natarajan, Sayee and Pickett, Larry A and Dutt, Varun},
  journal={Frontiers in big data},
  volume={3},
  pages={4},
  year={2020},
  publisher={Frontiers Media SA}
}

@article{lippi2013short,
  title={Short-term traffic flow forecasting: An experimental comparison of time-series analysis and supervised learning},
  author={Lippi, Marco and Bertini, Matteo and Frasconi, Paolo},
  journal={IEEE Transactions on Intelligent Transportation Systems},
  volume={14},
  number={2},
  pages={871--882},
  year={2013},
  publisher={IEEE}
}

@article{wei2022chain,
  title={Chain-of-thought prompting elicits reasoning in large language models},
  author={Wei, Jason and Wang, Xuezhi and Schuurmans, Dale and Bosma, Maarten and Xia, Fei and Chi, Ed and Le, Quoc V and Zhou, Denny and others},
  journal={Advances in neural information processing systems},
  volume={35},
  pages={24824--24837},
  year={2022}
}

@book{hyndman2018forecasting,
  title={Forecasting: principles and practice},
  author={Hyndman, Rob J and Athanasopoulos, George},
  year={2018},
  publisher={OTexts}
}

@misc{shumway2006time,
  title={Time Series Analysis and Its Applications: With R Examples},
  author={Shumway, Rober H},
  year={2006},
  publisher={Springer}
}

@article{ansari2024chronos,
  title={Chronos: Learning the language of time series},
  author={Ansari, Abdul Fatir and Stella, Lorenzo and Turkmen, Caner and Zhang, Xiyuan and Mercado, Pedro and Shen, Huibin and Shchur, Oleksandr and Rangapuram, Syama Sundar and Arango, Sebastian Pineda and Kapoor, Shubham and others},
  journal={arXiv preprint arXiv:2403.07815},
  year={2024}
}

@article{yang2025qwen3,
  title={Qwen3 technical report},
  author={Yang, An and Li, Anfeng and Yang, Baosong and Zhang, Beichen and Hui, Binyuan and Zheng, Bo and Yu, Bowen and Gao, Chang and Huang, Chengen and Lv, Chenxu and others},
  journal={arXiv preprint arXiv:2505.09388},
  year={2025}
}

@misc{bai2025qwen3vltechnicalreport,
      title={Qwen3-VL Technical Report}, 
      author={Shuai Bai and Yuxuan Cai and Ruizhe Chen and Keqin Chen and Xionghui Chen and Zesen Cheng and Lianghao Deng and Wei Ding and Chang Gao and Chunjiang Ge and Wenbin Ge and Zhifang Guo and Qidong Huang and Jie Huang and Fei Huang and Binyuan Hui and Shutong Jiang and Zhaohai Li and Mingsheng Li and Mei Li and Kaixin Li and Zicheng Lin and Junyang Lin and Xuejing Liu and Jiawei Liu and Chenglong Liu and Yang Liu and Dayiheng Liu and Shixuan Liu and Dunjie Lu and Ruilin Luo and Chenxu Lv and Rui Men and Lingchen Meng and Xuancheng Ren and Xingzhang Ren and Sibo Song and Yuchong Sun and Jun Tang and Jianhong Tu and Jianqiang Wan and Peng Wang and Pengfei Wang and Qiuyue Wang and Yuxuan Wang and Tianbao Xie and Yiheng Xu and Haiyang Xu and Jin Xu and Zhibo Yang and Mingkun Yang and Jianxin Yang and An Yang and Bowen Yu and Fei Zhang and Hang Zhang and Xi Zhang and Bo Zheng and Humen Zhong and Jingren Zhou and Fan Zhou and Jing Zhou and Yuanzhi Zhu and Ke Zhu},
      year={2025},
      eprint={2511.21631},
      archivePrefix={arXiv},
      primaryClass={cs.CV},
      url={https://arxiv.org/abs/2511.21631}, 
}

@article{guo2025deepseek,
  title={Deepseek-r1: Incentivizing reasoning capability in llms via reinforcement learning},
  author={Guo, Daya and Yang, Dejian and Zhang, Haowei and Song, Junxiao and Zhang, Ruoyu and Xu, Runxin and Zhu, Qihao and Ma, Shirong and Wang, Peiyi and Bi, Xiao and others},
  journal={arXiv preprint arXiv:2501.12948},
  year={2025}
}

@article{brown2020language,
  title={Language models are few-shot learners},
  author={Brown, Tom and Mann, Benjamin and Ryder, Nick and Subbiah, Melanie and Kaplan, Jared D and Dhariwal, Prafulla and Neelakantan, Arvind and Shyam, Pranav and Sastry, Girish and Askell, Amanda and others},
  journal={Advances in neural information processing systems},
  volume={33},
  pages={1877--1901},
  year={2020}
}

@inproceedings{devlin2019bert,
  title={Bert: Pre-training of deep bidirectional transformers for language understanding},
  author={Devlin, Jacob and Chang, Ming-Wei and Lee, Kenton and Toutanova, Kristina},
  booktitle={Proceedings of the 2019 conference of the North American chapter of the association for computational linguistics: human language technologies, volume 1 (long and short papers)},
  pages={4171--4186},
  year={2019}
}

@inproceedings{merrill2024language,
  title={Language models still struggle to zero-shot reason about time series},
  author={Merrill, Mike A and Tan, Mingtian and Gupta, Vinayak and Hartvigsen, Thomas and Althoff, Tim},
  booktitle={Findings of the Association for Computational Linguistics: EMNLP 2024},
  pages={3512--3533},
  year={2024}
}

@article{raffel2020exploring,
  title={Exploring the limits of transfer learning with a unified text-to-text transformer},
  author={Raffel, Colin and Shazeer, Noam and Roberts, Adam and Lee, Katherine and Narang, Sharan and Matena, Michael and Zhou, Yanqi and Li, Wei and Liu, Peter J},
  journal={Journal of machine learning research},
  volume={21},
  number={140},
  pages={1--67},
  year={2020}
}

@misc{no_robots,
  author = {Nazneen Rajani and Lewis Tunstall and Edward Beeching and Nathan Lambert and Alexander M. Rush and Thomas Wolf},
  title = {No Robots},
  year = {2023},
  publisher = {Hugging Face},
  journal = {Hugging Face repository},
  howpublished = {\url{https://huggingface.co/datasets/HuggingFaceH4/no_robots}}
}

@article{tu2025survey,
  title={A Survey on LLM Mid-Training},
  author={Tu, Chengying and Zhang, Xuemiao and Weng, Rongxiang and Li, Rumei and Zhang, Chen and Bai, Yang and Yan, Hongfei and Wang, Jingang and Cai, Xunliang},
  journal={arXiv preprint arXiv:2510.23081},
  year={2025}
}
\bibliographystyle{icml2026}

\newpage
\appendix
\onecolumn

\section{Downstream Tasks}

We evaluate our model on two prominent multi-task time series question answering (QA) benchmarks: Time-MQA \cite{kong2025time} and ChatTime \cite{wang2025chattime}. These benchmarks cover a wide range of time series analysis tasks, spanning both traditional numerical analysis and more advanced reasoning tasks. The datasets include various domains, such as healthcare, finance, energy, transportation, and more, providing a comprehensive foundation for model evaluation. The following sections describe the details of these benchmarks and their associated tasks.

\subsection{Time-MQA Benchmark}

The Time-MQA framework is a multi-task time series QA system that integrates natural language queries across several time series tasks. It contains approximately 200,000 question-answering pairs derived from 12 diverse domains, including healthcare, energy, finance, traffic, IoT, and the web. This extensive dataset is designed not only to support traditional numerical tasks, such as forecasting and anomaly detection, but also to handle more complex reasoning tasks through text-enhanced queries. The multi-task structure of Time-MQA makes it an ideal benchmark for evaluating the versatility and reasoning ability of models across a range of time series problems.

The specific tasks within the Time-MQA dataset are as follows:

\paragraph{Forecasting}
The forecasting task in Time-MQA involves predicting future time points based on known time series data. This task spans multiple domains, such as healthcare and energy, and requires models to accurately predict future trends from historical observations. For instance, forecasting heart rate trends or energy consumption based on previous data points.

\paragraph{Imputation}
The imputation task addresses the challenge of filling in missing values within time series data. Time series data often suffer from gaps due to various reasons such as sensor malfunctions or incomplete data collection. The imputation task requires models to predict the missing values, thus restoring the continuity of the time series. This is especially important in fields like healthcare, where data integrity is critical.

\paragraph{Anomaly Detection}
Anomaly detection is essential in identifying unusual patterns or deviations in time series data. This task requires models to detect outliers or anomalous events that deviate from expected behavior. In industrial and financial settings, detecting anomalies in real-time data can signal critical events, such as equipment failure or market disruptions.

\paragraph{Classification}
The classification task involves categorizing time series data into predefined classes. For example, sensor data from wearable devices could be classified into different human activities, such as walking, sitting, or jogging. This task tests the model's ability to recognize patterns in time series and assign them to appropriate categories.

\paragraph{True/False Questions}
The True/False questions assess the model's ability to verify simple assertions based on the time series data. These questions ask the model to determine whether a statement about the time series is correct or not.

\paragraph{Multiple Choice Questions}
In the Multiple Choice task, the model is asked to choose the correct answer from a set of predefined options. This task evaluates the model's ability to classify and compare specific characteristics of the time series data.

\paragraph{Open-Ended QA}
Open-ended QA tasks in Time-MQA require the model to generate detailed textual explanations about the time series data. These tasks go beyond simple numerical predictions, asking the model to interpret trends, identify causal relationships, and provide insights based on the time series patterns.

These tasks collectively make Time-MQA a comprehensive benchmark for evaluating multi-task models. By integrating numerical and textual reasoning, Time-MQA challenges models to provide more advanced insights into time series data, bridging the gap between classical numerical analysis and modern language model-driven approaches.

\subsection{ChatTime Benchmark}

The ChatTime framework is another state-of-the-art multimodal time series foundation model that incorporates both time series data and textual information. ChatTime uniquely handles both bimodal inputs and outputs, allowing models to leverage textual context alongside time series data. It offers zero-shot time series forecasting capabilities and facilitates multimodal time series question answering tasks, making it particularly useful for testing models on complex, context-sensitive tasks.

The ChatTime dataset includes tasks specifically designed to evaluate the model’s ability to process time series data alongside contextual textual information. The four main tasks within this dataset are as follows:

\paragraph{Trend Analysis}
This task involves identifying the overall direction of a given time series. The model is required to determine whether the data is exhibiting an increasing, decreasing, or stable trend over time. The objective is to recognize the general movement in the data points and categorize the series accordingly.

\paragraph{Volatility Analysis}
Volatility analysis focuses on assessing the extent of fluctuations within the time series. The model is tasked with identifying periods of high volatility, where the data points show significant variations, or periods of low volatility, where the data remains relatively stable. This task requires recognizing how the data's variation changes over time.

\paragraph{Seasonality Analysis}
In seasonality analysis, the model is asked to determine if the time series displays periodic or cyclic behavior. This includes recognizing repeating patterns or trends that occur at regular intervals, indicating a seasonal influence on the data over time.

\paragraph{Outlier Detection}
Outlier detection requires the model to identify data points in the time series that significantly deviate from the general pattern or trend. These outliers might represent unusual events or anomalies, and the model needs to pinpoint these points based on their departure from expected behavior.

Each of these tasks challenges the model to integrate both time series data and textual information, requiring not only numerical analysis but also a deeper understanding of the context within which the data is generated. The multimodal nature of ChatTime makes it an ideal candidate for evaluating models on their ability to reason across both domains simultaneously.

\section{Implementation Details of Evaluation}

We provide the implementation details for our evaluation. For the ChatTime benchmark, given the massive scale of synthesized time series question-answering pairs, we randomly sample 300 instances per task as the test set to evaluate the accuracy of LLMs in identifying time series patterns. For the Time-MQA benchmark, we follow the experimental settings in the original text by extracting 50 samples per task as test cases.Specifically, for forecasting and imputation tasks, we employ the Symmetric Mean Absolute Percentage Error (SMAPE) as the evaluation metric. This choice is justified by the fact that the dataset spans multiple application domains with extreme numerical variances and diverse sequence stationarity; $\text{SMAPE}$ effectively assesses the quality of generation across these heterogeneous samples. The formula is defined as follows:
\begin{equation}
\text{SMAPE} = \frac{1}{n} \sum_{t=1}^{n} \frac{|F_t - A_t|}{(|A_t| + |F_t|)/2}
\end{equation}
where $A_t$ represents the actual value and $F_t$ denotes the forecast value. In rare cases where the sequence length generated by the LLM deviates from the instructions, we introduce a penalty factor of $1.1$. The adjusted error, $E_{adj}$, is calculated as:
\begin{equation}
E_{adj} = 
\begin{cases} 
E \times 1.1, & \text{if } L_{gen} \neq L_{inst} \\
E, & \text{if } L_{gen} = L_{inst}
\end{cases}
\end{equation}
where $L_{inst}$ and $L_{gen}$ refer to the instructed and generated lengths, respectively. Except for open-ended questions, all other tasks utilize Accuracy (ACC) as the evaluation metric. For the open-ended task in Time-MQA and the KnoTS benchmark, we use Gemini-3-Pro-Preview for scoring. The model provides a comprehensive score by considering multiple dimensions, including factual correctness, temporal logic, completeness, and conciseness.

\section{Error Bar Analysis}\label{app:error-bar}

To ensure the reproducibility and statistical reliability of our results, we conducted three independent runs ($N=3$) for our main experiments using different random seeds. The values are reported as \textbf{mean $\pm$ standard deviation}, which accounts for the inherent stochasticity in LLM decoding and sampling processes. This analysis confirms that the performance gains observed in our study are consistent across multiple iterations and are not a result of favorable random seeding, demonstrating the stability of our proposed method.

\begin{table*}[h]
\caption{Robustness of Thoth-30B-A3B and Thoth-8B performance on all benchmarks. The results are obtained from three random runs.}
\label{tab:error-bar}
\centering
\resizebox{\textwidth}{!}{
\begin{tabular}{l|cccc | ccccccc | cc}
\toprule
Model & \multicolumn{4}{c|}{ChatTime Benchmark} & \multicolumn{7}{c|}{Time-MQA Benchmark} & \multicolumn{2}{c}{KnoTS Benchmark}\\ \midrule
& Trend $\uparrow$ & Volatility $\uparrow$ & Seasonality $\uparrow$ & Outliers $\uparrow$ & Forecast $\downarrow$ & Impu. $\downarrow$ & Clas. $\uparrow$ & Anom. $\uparrow$ & Judge. $\uparrow$ & MCQ $\uparrow$ & Open. $\uparrow$ & Reason $\uparrow$ & Decision $\uparrow$ \\
\midrule
Thoth-8B & 0.973\scalebox{0.9}{$\pm$0.003} & 0.750\scalebox{0.9}{$\pm$0.005} & 0.603\scalebox{0.9}{$\pm$0.007} & 0.753\scalebox{0.9}{$\pm$0.003} & 0.522\scalebox{0.9}{$\pm$0.019} & 0.333\scalebox{0.9}{$\pm$0.013} & 0.10\scalebox{0.9}{$\pm$0.00} & 0.58\scalebox{0.9}{$\pm$0.01} & 0.72\scalebox{0.9}{$\pm$0.02} & 0.64\scalebox{0.9}{$\pm$0.02} & 5.98\scalebox{0.9}{$\pm$0.17} & 5.54\scalebox{0.9}{$\pm$0.24} & 3.21\scalebox{0.9}{$\pm$0.18} \\
Thoth-30B-A3B & 0.957\scalebox{0.9}{$\pm$0.003} & 0.770\scalebox{0.9}{$\pm$0.010} & 0.573\scalebox{0.9}{$\pm$0.010} & 0.740\scalebox{0.9}{$\pm$0.007} & 0.432\scalebox{0.9}{$\pm$0.021} & 0.247\scalebox{0.9}{$\pm$0.014} & 0.02\scalebox{0.9}{$\pm$0.00} & 0.70\scalebox{0.9}{$\pm$0.00} & 0.78\scalebox{0.9}{$\pm$0.01} & 0.72\scalebox{0.9}{$\pm$0.02} & 7.42\scalebox{0.9}{$\pm$0.11} & 8.05\scalebox{0.9}{$\pm$0.21} & 4.96\scalebox{0.9}{$\pm$0.31} \\
\bottomrule
\end{tabular}
}
\end{table*}

\section{Prompts of \emph{Book-of-Thoth} Generation}

\begin{figure}[ht]
  \begin{center}
    \centerline{\includegraphics[width=1\columnwidth]{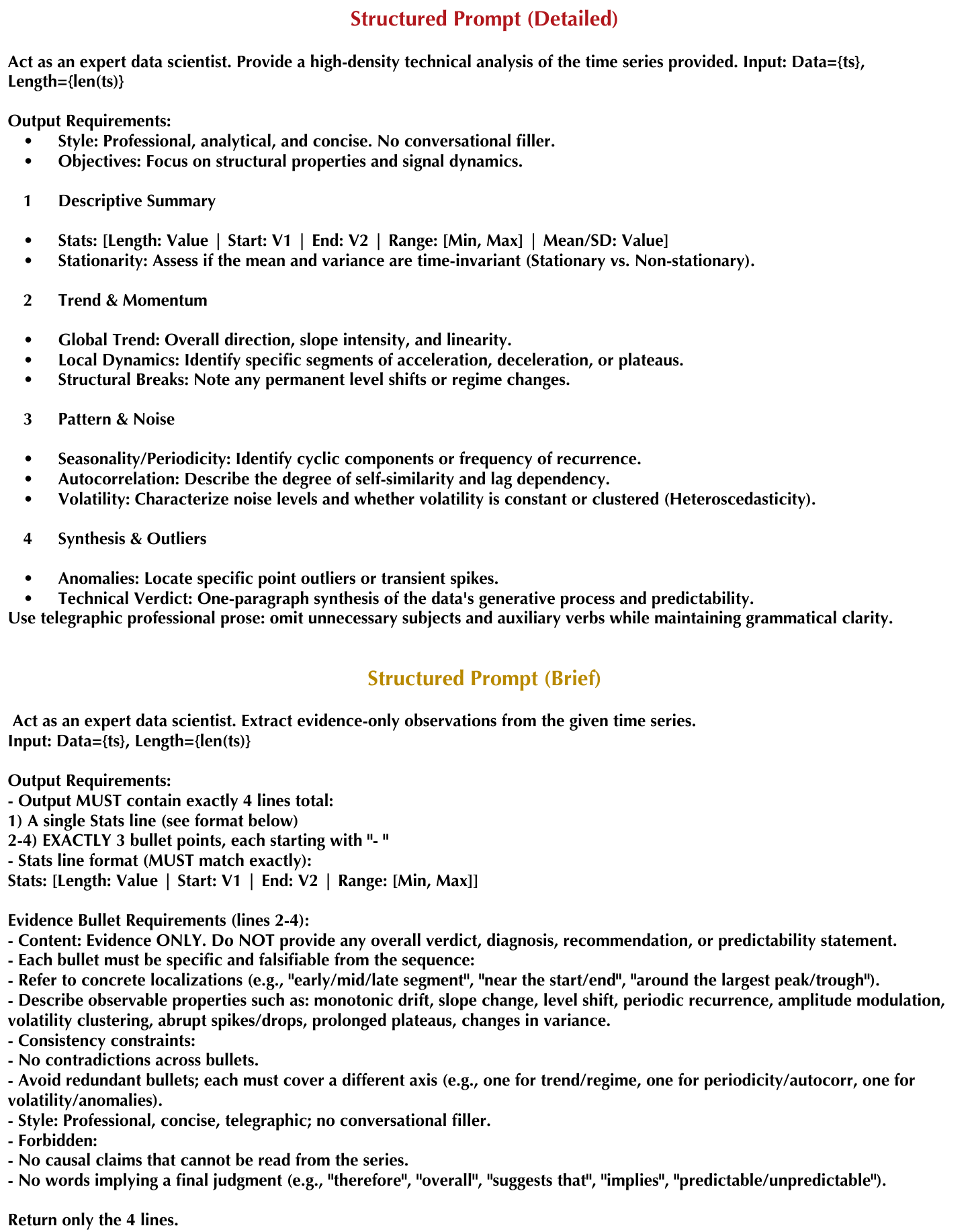}}
    \caption{
    Examples of carefully designed prompts used for constructing structured data of \emph{Book-of-Thoth}.
    }
    \label{fig:prompt1}
  \end{center}
\end{figure}

\begin{figure}[ht]
  \begin{center}
    \centerline{\includegraphics[width=1\columnwidth]{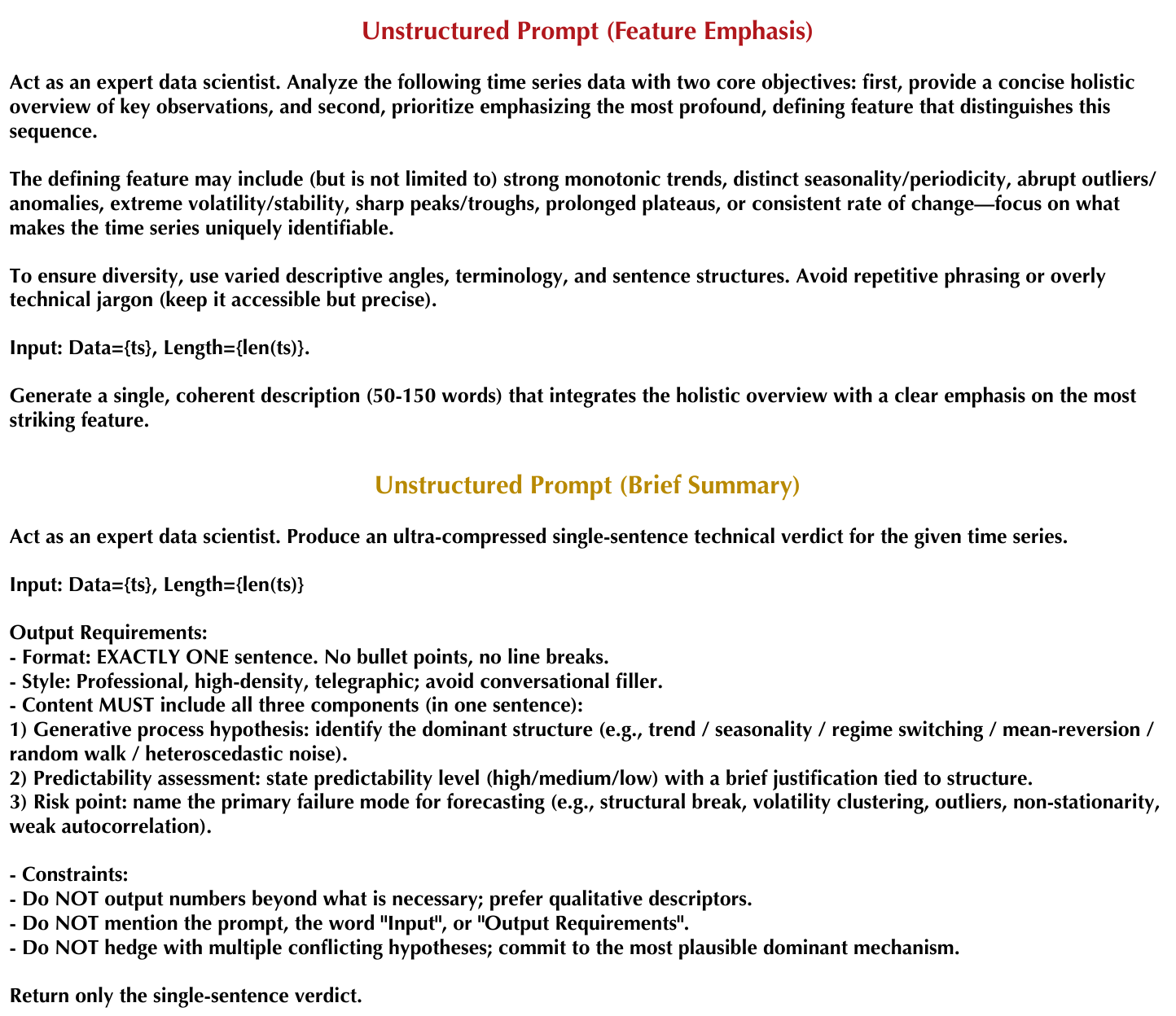}}
    \caption{
      Examples of carefully designed prompts used for constructing unstructured data of \emph{Book-of-Thoth}.
    }
    \label{fig:prompt2}
  \end{center}
\end{figure}

In this section, we present the details for constructing the mid-training general-purpose dataset, \emph{Book-of-Thoth}. For the diverse time series data synthesized via the KernelSynth method, we design a set of carefully crafted prompts to elicit high-quality and diverse time series–text pairs from GPT-5.2. For both structured and unstructured text, we design two versions of prompts, which generate detailed and brief texts, respectively. This significantly enhances the diversity of the generated data. The prompting scheme and examples are shown in Figure \ref{fig:prompt1} and Figure \ref{fig:prompt2}. 

\section{Details of KnoTS} \label{app:knots-cases}

KnoTS is designed to better evaluate time series reasoning and decision-making capabilities of LLMs under knowledge-intensive settings. To assess the linguistic complexity and diversity of existing time series benchmarks, we conduct a comprehensive lexical analysis, with the results presented in Table \ref{tab:lexical-analysis}. Lexical richness is quantified using the Type-Token Ratio ($TTR$), and structural variety is quantified using N-gram entropy ($H_n$):

\begin{equation}
    TTR = \frac{|V|}{N}, \quad H_n = -\sum_{w \in V_n} P(w) logP(w)
\end{equation}

Specifically, $TTR$ is the ratio of unique vocabulary size $|V|$ to total token count $N$, while $H_n$ is calculated from the probability $P(w)$ of each unique n-gram $w$ within the set $V_n$. Our analysis, performs on a uniformly sampled subset, reveals a substantial gap between the proposed KnoTS benchmark and prior datasets. KnoTS exhibits a markedly higher type–token ratio and consistently elevated n‑gram entropy scores, quantitatively confirming that it has a richer vocabulary and a more diverse linguistic structure. This advantage stems from our construction methodology, which eschews the rigid templates common in other datasets.

\begin{table}[ht]
\caption{Diversity and complexity analysis of time series-centric questions in different benchmarks. }
\label{tab:lexical-analysis}
\begin{center}
\begin{small}
\begin{sc}
\resizebox{0.6\columnwidth}{!}{
\begin{tabular}{l|ccccc}
\toprule
Dataset & TTR $\uparrow$ & 1-Ent $\uparrow$ & 2-Ent $\uparrow$ & 3-Ent $\uparrow$ & Avg-Len \\
\midrule
\textbf{KnoTS (Ours)} & \textbf{0.2109} & \textbf{9.69} & \textbf{13.02} & \textbf{13.67} & 169.8 \\
MTBench \citeyearpar{chen2025mtbench} & 0.1131 & 7.54 & 10.69 & 12.00 & 77.5 \\
Time-MQA \citeyearpar{kong2025time} & 0.0877 & 7.35 & 8.90 & 9.25 & 49.7 \\
ChatTime \citeyearpar{wang2025chattime} & 0.0098 & 5.56 & 6.86 & 7.40 & 90.4 \\
ChatTS \citeyearpar{xie2024chatts} & 0.0309 & 7.16 & 8.89 & 9.45 & 293.7 \\
TS-Insights \citeyearpar{zhang2025insight} & 0.0118 & 4.81 & 5.05 & 5.10 & 37.0 \\
\bottomrule
\end{tabular}
}
\end{sc}
\end{small}
\end{center}
\vskip -0.1in
\end{table}

\begin{figure}[ht]
  \begin{center}
    \centerline{\includegraphics[width=0.85\columnwidth]{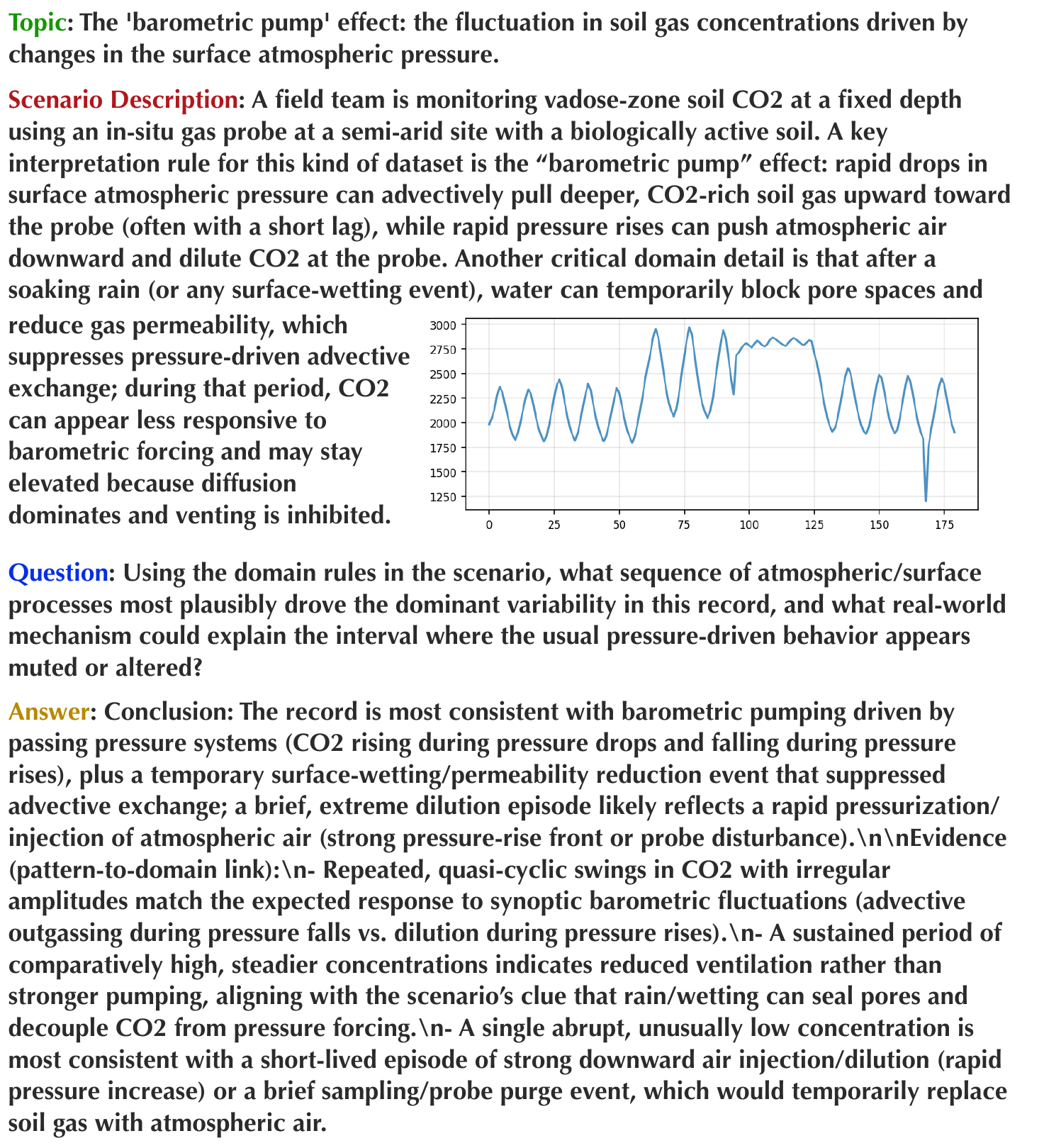}}
    \caption{
      A complete example of reasoning questions in the KnoTS benchmark. It is the full version of Figure \ref{KnoTS-brief-case} in the main text.
    }
    \label{fig:knots-case2}
  \end{center}
\end{figure}

\begin{figure}[ht]
  \begin{center}
    \centerline{\includegraphics[width=0.85\columnwidth]{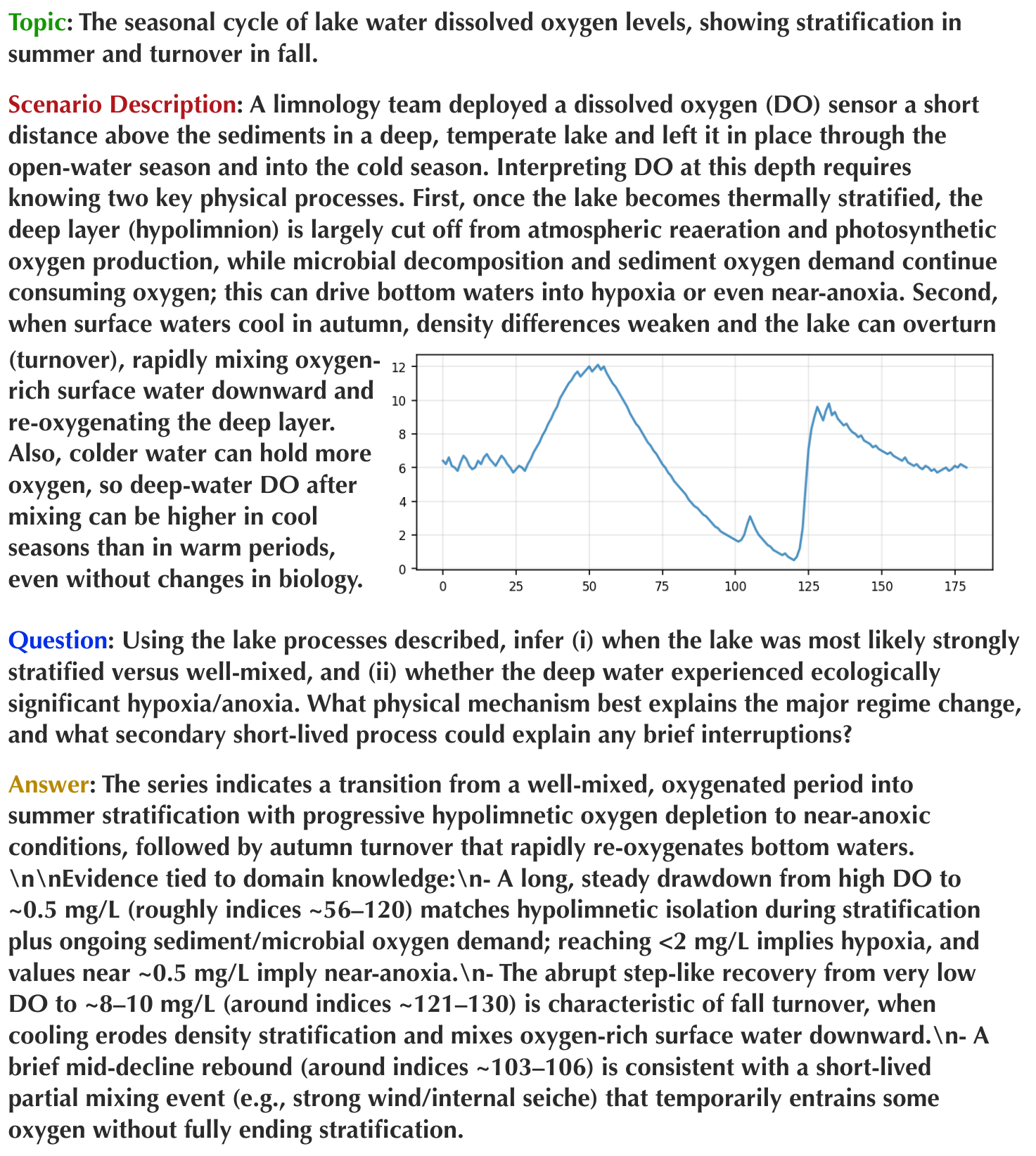}}
    \caption{
      A complete example of reasoning questions in the KnoTS benchmark.
    }
    \label{fig:knots-case1}
  \end{center}
\end{figure}

\begin{figure}[ht]
  \begin{center}
    \centerline{\includegraphics[width=0.85\columnwidth]{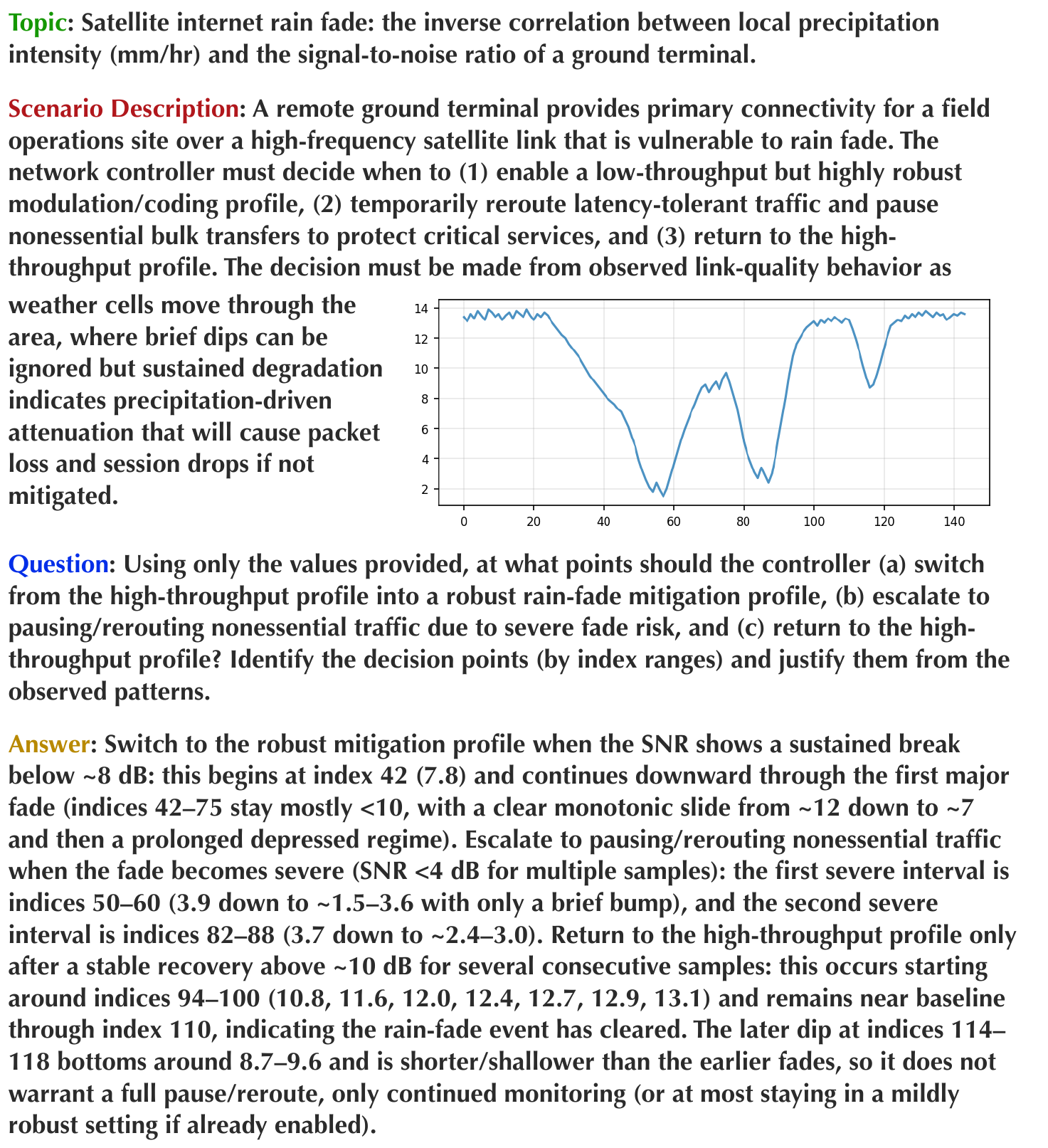}}
    \caption{
      A complete example of decision-making questions in the KnoTS benchmark.
    }
    \label{fig:knots-case3}
  \end{center}
\end{figure}

Further, we present two representative reasoning examples and one decision-making example in Figures \ref{fig:knots-case2}, Figure \ref{fig:knots-case1}, Figure \ref{fig:knots-case3}. These instances share two key characteristics: they are tightly grounded in domain background knowledge, and they require models to reason and act based on the concrete time series evidence. As such, KnoTS enables evaluation of time series understanding while also testing whether incorporating time series compromises the model’s general reasoning and decision-making competence.

\section{Showcases}\label{app:showcases}

\begin{figure}[ht]
  \begin{center}
    \centerline{\includegraphics[width=0.7\columnwidth]{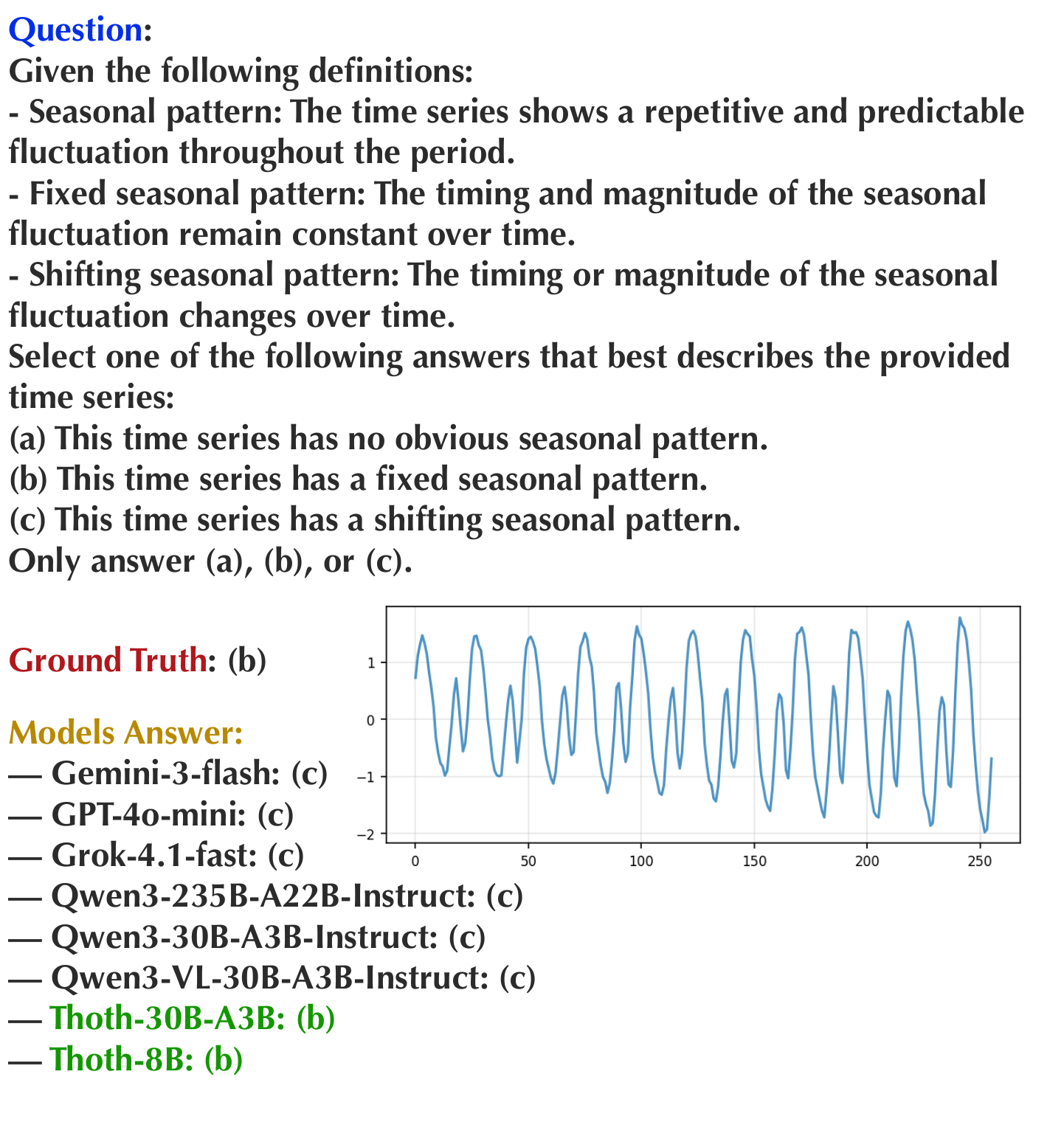}}
    \caption{
      A showcase of seasonality recognition question from various LLMs in the ChatTime benchmark.
    }
    \label{fig:showcase1}
  \end{center}
\end{figure}

\begin{figure}[ht]
  \begin{center}
    \centerline{\includegraphics[width=0.7\columnwidth]{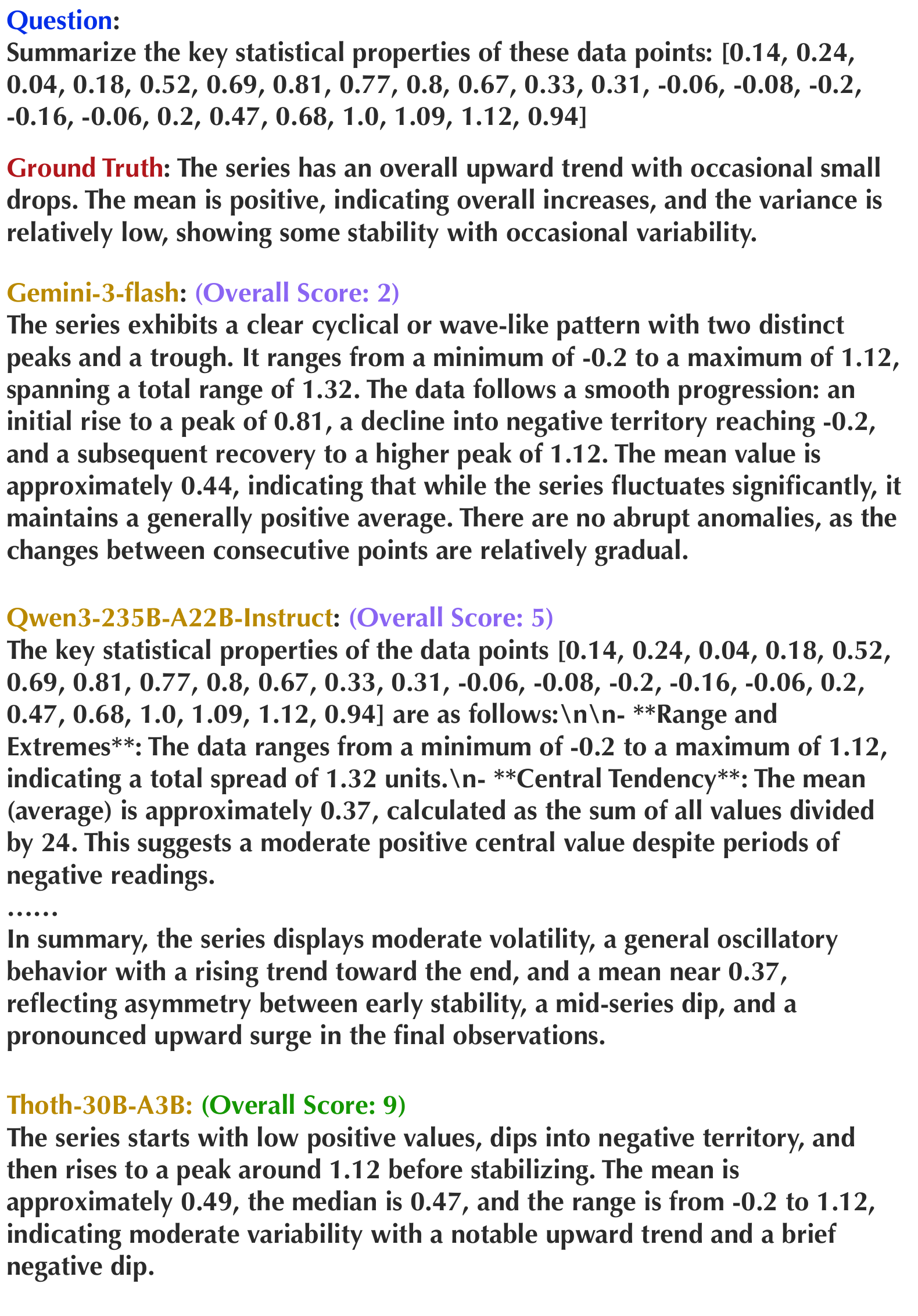}}
    \caption{
    A showcase of an open-ended question from three LLMs in the Time-MQA benchmark.
    }
    \label{fig:showcase2}
  \end{center}
\end{figure}

\begin{figure}[ht]
  \begin{center}
    \centerline{\includegraphics[width=0.92\columnwidth]{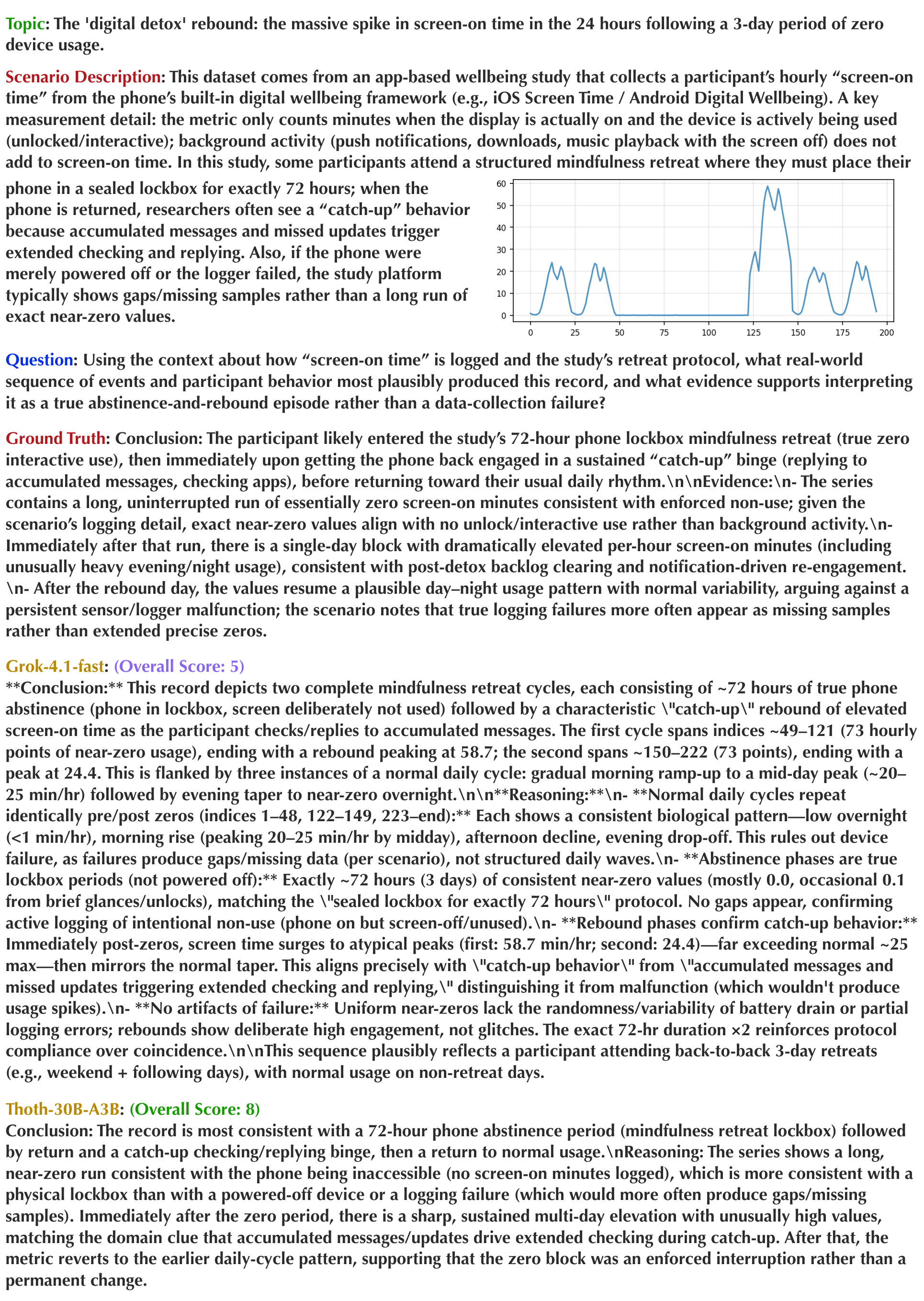}}
    \caption{
     A showcase of reasoning question from Grok-4.1-Fast and Thoth-30B-A3B in the KnoTS benchmark.
    }
    \label{fig:showcase3}
  \end{center}
\end{figure}

In this section, we present three complete examples of model inference, as illustrated in Figures \ref{fig:showcase1}, \ref{fig:showcase2}, and \ref{fig:showcase3}. For each benchmark, we select one representative problem to compare the inference results of Thoth with the baseline LLMs.

\section{Full Results}\label{app:chattime-full-results}

The ChatTime benchmark categorizes input sequences into four lengths: 64, 128, 256, and 512. However, since the 512 length sequences significantly challenge the input capacity of LLMs under n-shot in-context learning settings, we restrict our evaluation to lengths of 64, 128, and 256. The complete results are detailed in Table \ref{tab:full-results}. Both parameter versions of Thoth demonstrate robust performance across all tested input lengths.

\begin{table*}[h]
\caption{Full results of ChatTime benchmark \cite{wang2025chattime}. There are three fixed time series lengths of 64, 128, and 256 of the four time series pattern recognition tasks. \emph{Avg} means the average results from the three lengths.}
\label{tab:full-results}
\centering
\resizebox{\textwidth}{!}{
\begin{tabular}{l | cccc | cccc | cccc | cccc}
\toprule
Model & \multicolumn{4}{c|}{Trend}  & \multicolumn{4}{c|}{Volatility}  & \multicolumn{4}{c|}{Seasonality}  & \multicolumn{4}{c}{Outliers} \\
& 64 & 128 & 256 & \emph{Avg} & 64 & 128 & 256 & \emph{Avg} & 64 & 128 & 256 & \emph{Avg} & 64 & 128 & 256 & \emph{Avg} \\
\midrule
\multicolumn{17}{c}{\cellcolor{myblue}\textit{Proprietary Models}} \\
\midrule
Gemini-3-Flash-Preview & \secondres{0.970} & 0.800 & 0.840 & 0.870 & \boldres{0.880} & \boldres{0.860} & \boldres{0.920} & \boldres{0.887} & 0.550 & 0.550 & 0.460 & 0.520 & 0.700 & \boldres{0.800} & \secondres{0.740} & \secondres{0.747} \\
GPT-4o-mini & 0.890 & 0.790 & \secondres{0.870} & 0.850 & 0.380 & 0.500 & 0.420 & 0.433 & 0.540 & 0.370 & 0.370 & 0.427 & 0.660 & 0.660 & 0.700 & 0.673 \\
Grok-4.1-Fast & 0.680 & 0.700 & 0.560 & 0.647 & 0.460 & 0.380 & 0.310 & 0.383 & 0.590 & \boldres{0.730} & 0.540 & \boldres{0.620} & 0.700 & 0.750 & \boldres{0.790} & \secondres{0.747} \\
\midrule
\multicolumn{17}{c}{\cellcolor{myred}\textit{Open-source Large Language Models}} \\
\midrule
Qwen3-235B-A22B-Instruct & 0.940 & 0.820 & 0.840 & 0.867 & 0.500 & 0.560 & 0.480 & 0.513 & 0.550 & 0.350 & 0.440 & 0.447 & 0.710 & 0.650 & 0.640 & 0.667 \\
Qwen3-30B-A3B-Instruct & 0.840 & 0.700 & 0.800 & 0.780 & 0.490 & 0.480 & 0.430 & 0.467 & 0.400 & 0.310 & 0.300 & 0.337 & 0.530 & 0.450 & 0.350 & 0.443 \\
Deepseek-R1-32B & 0.540 & 0.670 & 0.390 & 0.533 & 0.530 & 0.460 & 0.360 & 0.450 & 0.510 & 0.350 & 0.480 & 0.447 & 0.720 & 0.640 & 0.510 & 0.623 \\
Mistral-Small-24B-Instruct & 0.860 & 0.870 & 0.820 & 0.850 & 0.530 & 0.550 & 0.460 & 0.513 & 0.380 & 0.340 & 0.350 & 0.357 & 0.690 & 0.580 & 0.610 & 0.627 \\
Llama-3.1-8B-Instruct & 0.800 & 0.410 & 0.290 & 0.500 & 0.360 & 0.380 & 0.310 & 0.350 & 0.250 & 0.400 & 0.300 & 0.317 & 0.360 & 0.410 & 0.380 & 0.383 \\
Qwen3-8B & 0.380 & 0.050 & 0.060 & 0.163 & 0.310 & 0.340 & 0.010 & 0.220 & 0.270 & 0.020 & 0.010 & 0.100 & 0.710 & 0.610 & 0.020 & 0.447 \\
\midrule
\multicolumn{17}{c}{\cellcolor{mypurple}\textit{Open-source Vision Language Models}} \\
\midrule
Qwen3-VL-30B-A3B-Instruct & 0.920 & 0.810 & 0.850 & 0.860 & 0.230 & 0.510 & 0.640 & 0.460 & 0.440 & 0.250 & 0.360 & 0.350 & 0.480 & 0.490 & 0.440 & 0.470 \\
Mistral-Small-3.2-24B-Instruct & 0.460 & 0.500 & 0.330 & 0.430 & 0.410 & 0.380 & 0.330 & 0.373 & 0.530 & \secondres{0.620} & \boldres{0.640} & 0.597 & 0.610 & 0.570 & 0.510 & 0.563 \\
\midrule
\rowcolor{gray!20} \multicolumn{17}{c}{\textit{Task-Specific Time Series Language Models}} \\
\midrule
\rowcolor{gray!20} ChatTime-7B-Chat & 0.990 & 0.980 & 0.970 & 0.980 & 0.820 & 0.910 & 0.880 & 0.870 & 0.760 & 0.720 & 0.710 & 0.730 & 0.980 & 0.990 & 0.920 & 0.963 \\
\rowcolor{gray!20} Time-MQA(LLaMA3-8B) & 0.720 & 0.580 & 0.660 & 0.653 & 0.410 & 0.500 & 0.390 & 0.433 & 0.330 & 0.230 & 0.340 & 0.300 & 0.480 & 0.360 & 0.340 & 0.393 \\
\rowcolor{gray!20} Time-MQA(Qwen2.5-7B) & 0.270 & 0.390 & 0.230 & 0.297 & 0.350 & 0.410 & 0.350 & 0.370 & 0.340 & 0.400 & 0.290 & 0.343 & 0.350 & 0.370 & 0.450 & 0.390 \\
\midrule
\multicolumn{17}{c}{\cellcolor{mygreen}\textit{Ours}} \\
\midrule
\textbf{Thoth-8B} & \boldres{0.980} & \boldres{0.970} & \boldres{0.970} & \boldres{0.973} & \secondres{0.770} & 0.770 & 0.710 & 0.750 & \boldres{0.610} & 0.580 & \secondres{0.620} & \secondres{0.603} & \secondres{0.780} & \secondres{0.760} & 0.720 & \boldres{0.753}   \\
\textbf{Thoth-30B-A3B} & \boldres{0.980} & \secondres{0.920} & \boldres{0.970} & \secondres{0.957} & \secondres{0.770} & \secondres{0.790} & \secondres{0.750} & \secondres{0.770} & \secondres{0.600} & 0.520 & 0.600 & 0.573 & \boldres{0.820} & 0.720 & 0.680 & 0.740 \\
\bottomrule
\end{tabular}
}
\end{table*}


\end{document}